\RequirePackage{fix-cm} % Exact scalable math sizes in appendix diagrams.
\documentclass[10pt,twocolumn,letterpaper]{article}
\usepackage[pagenumbers]{cvpr}
\usepackage{graphicx}
\usepackage{array}
\usepackage{placeins}
\usepackage{tikz}
\usetikzlibrary{arrows.meta,positioning,calc}
\definecolor{cvprblue}{rgb}{0.21,0.49,0.74}
\definecolor{ink}{HTML}{233B4F}
\definecolor{bluewash}{HTML}{EBF3FA}
\definecolor{orangewash}{HTML}{FFF2E1}
\definecolor{greenwash}{HTML}{EAF6F0}
\definecolor{violetwash}{HTML}{F0ECFA}

\usepackage{pgfplots}
\pgfplotsset{compat=1.18}
\definecolor{nexInk}{HTML}{30363B}
\definecolor{nexBlue}{HTML}{50758F}
\definecolor{nexTeal}{HTML}{008C82}
\definecolor{nexOrange}{HTML}{C78B45}
\definecolor{nexPurple}{HTML}{8267AB}
\definecolor{nexRust}{HTML}{BD694F}
\pgfplotsset{nex axis/.style={
 scale only axis,axis background/.style={fill=white},
 axis line style={nexInk!40,line width=.5pt},
 tick style={nexInk!35,line width=.5pt},
 tick label style={font=\fontsize{8.5}{10}\selectfont,text=nexInk},
 label style={font=\fontsize{9}{11}\selectfont,text=nexInk},
 title style={font=\bfseries\fontsize{9.5}{11}\selectfont,text=nexInk},
 grid style={nexInk!10,line width=.3pt},axis on top,
 legend style={draw=none,fill=white,font=\fontsize{8.5}{10}\selectfont},
}}
\usepackage[pagebackref,breaklinks,colorlinks,allcolors=cvprblue]{hyperref}
\def\confName{CVPR}
\def\confYear{2027}
\hypersetup{pdftitle={NexSight: Verifiable Tasks for Fine-Grained Agentic Long-Video Understanding},pdfauthor={Tianci Xie, Li Chen, Pinxue Guo, Rui Zheng}}

\title{\raisebox{-0.30em}{\includegraphics[height=1.55em,trim=180 65 160 70,clip]{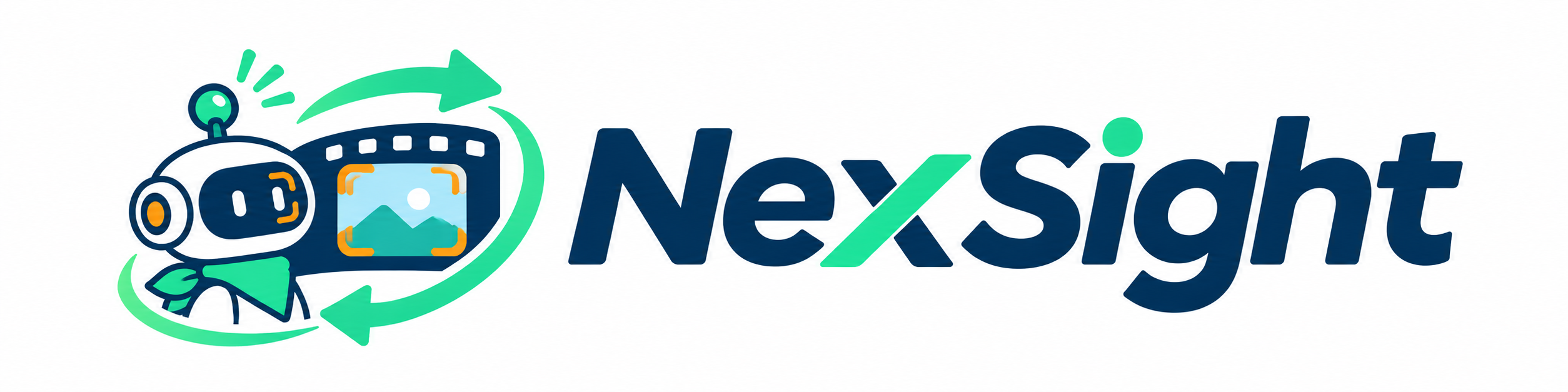}}: Verifiable Tasks for Fine-Grained\\Agentic Long-Video Understanding}
\author{Tianci Xie\textsuperscript{1}\qquad Li Chen\textsuperscript{2}\qquad Pinxue Guo\textsuperscript{3}\thanks{Corresponding author.}\qquad Rui Zheng\textsuperscript{3}\\
\textsuperscript{1}Zhejiang University\qquad \textsuperscript{2}Tongji University\\
\textsuperscript{3}Shanghai Qiji Zhifeng Co., Ltd.}
\begin{document}
\maketitle
\begin{abstract}
Fine-grained questions about long videos ask how many times an action occurs, in what order events unfold, or exactly when a state changes. Answering them can require finding brief moments scattered across hours of footage. Agentic models that decide when and where to look promise to find these moments, yet this ability is hard both to measure and to learn. Sparse supporting evidence may establish that an action occurred without establishing its total count; verifying timeline-wide claims requires checking for missed occurrences and double counting. Answer-only supervision, in turn, does not explicitly supervise the search for evidence. We present NexSight, which builds verifiable tasks to both evaluate and train agentic long-video understanding. Its agentic pipeline, SightForge, accepts a task only if a verifier blind to the draft answer can reconstruct it from evidence densely reacquired from the raw video and counterexample checks fail to refute it. This yields SightLine, 310 evidence-traceable tasks from 108 long videos, evaluated through SightLoop, a shared multiscale harness supporting temporal exploration, local inspection, evidence integration, and revisiting while recording returned observations. On disjoint videos, solver trajectories on verified tasks become step-level supervision for observation decisions. The strongest model reaches only 54.52\% on SightLine, yet 79.35\% when given the annotated evidence. With under 6K examples, Sight-35B nearly doubles its base model's interactive accuracy (21.29\%$\rightarrow$42.26\%), outperforms the 397B model of its family, begins using explicit spatial inspection after post-training, which its base model never uses, and gains over 8 points on two external long-video benchmarks.
\end{abstract}

\section{Introduction}
\label{sec:intro}

Long videos record how events unfold over minutes to hours, and many questions about them are fine-grained: how many times an action repeats, which of two distant events came first, or exactly when a state changes. The answers hinge on a few seconds scattered across the timeline, which uniform sampling, spreading a fixed frame budget thinly over hours, easily skips. Agentic long-video understanding instead lets the model decide when to search the timeline, which details to inspect, and how to integrate evidence across observations~\citep{ye2025temporal_search,wang2025avp,yang2025longvt,wang2024videoagent,wang2024videotree}, spending its budget where the question demands. Realizing this promise requires two things that remain open: \emph{measuring} whether an agent acquires the right evidence and uses it correctly, and \emph{teaching} it to do so.

Measuring requires tasks whose answers are correct and whose evidence suffices for anyone to reconstruct them. Fine-grained tasks make this hard in principle: many are \emph{global claims}, such as a count or an order, and proving that an action occurred exactly three times requires \emph{negative} evidence that no occurrence was missed or counted twice. Grounded benchmarks associate answers with supporting frames or intervals~\citep{xiao2024nextgqa,chen2025cgbench}, but sparse support alone may not suffice to verify timeline-wide claims, and a verifier that sees the draft answer tends to confirm rather than reconstruct it~\citep{dhuliawala2024cove}. With unverified references, errors cannot be attributed to missed evidence, faulty reasoning, or flawed tasks, and agentic evaluations, which mostly score answers or milestones~\citep{wang2025avp,an2026agentvidbench,yu2026longvidsearch} rather than the frames actually decoded~\citep{liu2026trace}, further entangle acquisition with use.

Teaching this ability is harder still: agents must decide where to look next, what to zoom into, and when the evidence suffices, yet final-answer supervision does not specify these decisions. Tool access is not enough: the evaluated instruction-tuned 35B model scores lower when allowed to search (28.06\%) than when simply given 64 uniform frames (33.23\%). Learning these decisions requires tasks that genuinely demand evidence seeking, answers trustworthy enough to supervise, and records of the search itself; few prior works train on their constructed data~\citep{yang2025longvt} (Table~\ref{tab:benchmark_comparison}).

Our key insight is that verification is itself an active-observation problem. Confirming a global claim means returning to the raw video, gathering dense evidence where the claim lives, and searching where counterexamples could hide. These are the same observation decisions we want agents to learn. A pipeline that annotates from a fixed view cannot do this; an agentic one can. The resulting verified tasks serve as both yardstick and training ground: their evidence certifies references and measures what agents observe, and on held-out videos they demand the very evidence seeking we aim to teach.

\begin{figure*}[t]
\centering
\begin{tikzpicture}
\node[anchor=south west,inner sep=0] (overview) at (0,0) {\includegraphics[width=\textwidth]{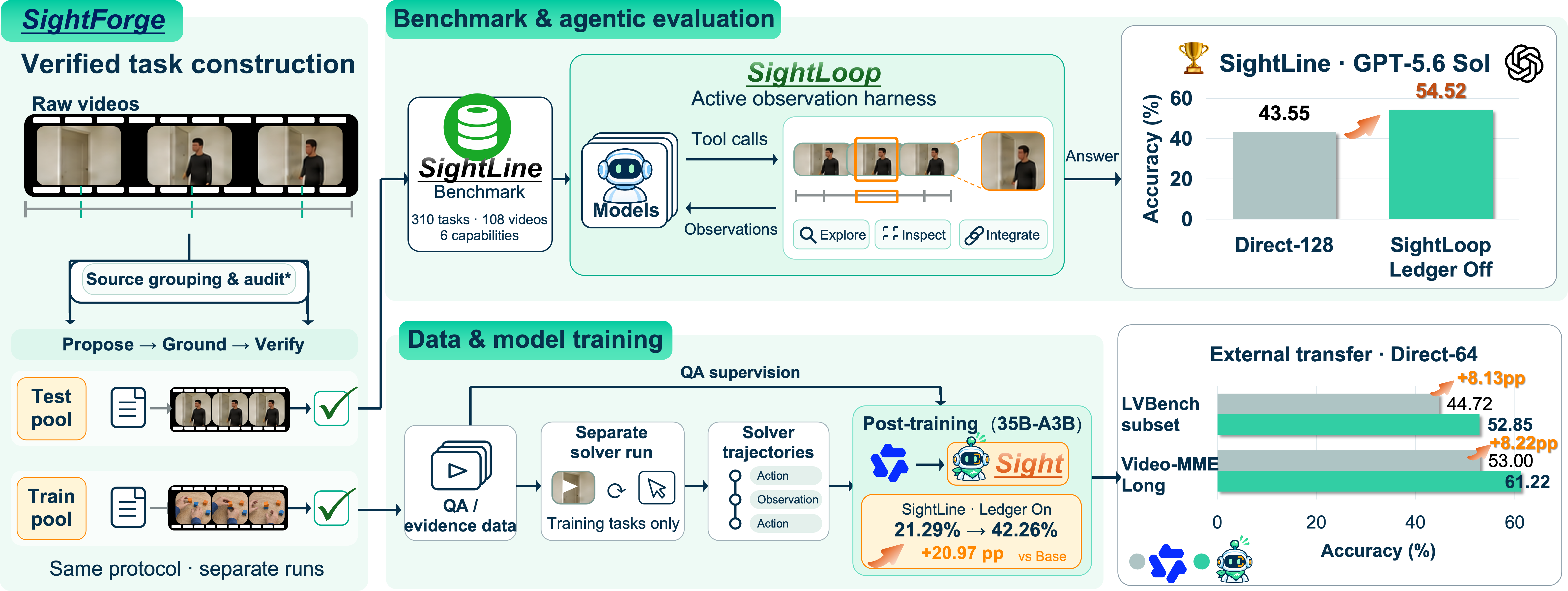}};
% Hide the obsolete audit-footnote marker without resampling the source image.
\begin{scope}[x={(overview.south east)},y={(overview.north west)}]
\definecolor{auditlabelbg}{RGB}{247,252,250}
\fill[auditlabelbg] (0.18130,0.52665) rectangle (0.18606,0.54050);
\end{scope}
\end{tikzpicture}
\caption{\textbf{NexSight overview.} (1) \textbf{Verifiable task construction (SightForge):} dense raw-video evidence, answer-hidden verification, and counterexample checks establish reliable tasks. (2) \textbf{Multiscale active evaluation (SightLine and SightLoop):} a 310-task benchmark and interactive harness assess temporal exploration, local inspection, and evidence integration. (3) \textbf{Agentic post-training (Sight-35B):} QA supervision and separately collected solver trajectories improve interactive accuracy by 20.97 percentage points over Base and support Direct-64 transfer to Video-MME Long and LVBench (123-question subset).}
\label{fig:motivation}
\end{figure*}

We build NexSight on this insight (Figure~\ref{fig:motivation}). SightForge constructs verifiable tasks for two paths: SightLine supports fine-grained evaluation through SightLoop's budgeted, multiscale observation interface; separately constructed training tasks provide QA supervision and support solver-trajectory collection. Solvers operate without reference answers or annotated evidence. Their interaction records complement QA supervision for post-training Sight-35B, connecting reliable task construction with both the evaluation and learning of active observation.

Evaluation shows that SightLine is far from solved: the strongest configuration reaches only 54.52\%, yet 79.35\% when given the annotated evidence, making acquisition a key bottleneck. At an identical frame budget, agentic window selection doubles temporal evidence recall over fixed windows (19.84\%$\rightarrow$40.63\%), yet cross-window counting and state change do not improve: finding the right moments is not enough without integrating them.

Training teaches this behavior. Sight-35B raises interactive accuracy from 21.29\% to 42.26\%, overtaking the instruction-tuned model by 14.19 points and every configuration of the 397B model of its family. Its largest gains come on cross-window counting (+37.50) and state change (+35.19), precisely where better window selection alone did not help, and it begins using explicit spatial inspection after post-training (0.72 calls per trajectory), which its base model never uses. The gains transfer: +8.22 points on Video-MME Long (+9.33 on training-disjoint videos) and +8.13 on an LVBench subset.

Our contributions are threefold:
\begin{enumerate}[label=(\arabic*),leftmargin=*,itemsep=2pt,topsep=3pt]
\item \textbf{Verifiable task construction.} \textbf{SightForge} separates task proposals from acceptance through dense raw-video evidence acquisition, answer-hidden verification, independent review, and counterexample checks, yielding \textbf{SightLine}, a six-capability benchmark with traceable visual evidence (Secs.~\ref{sec:benchmark} and~\ref{sec:construction}).

\item \textbf{Observation-level evaluation.} \textbf{SightLoop} couples temporal exploration, local inspection, evidence integration, and revisiting through a shared raw-video interface under explicit budgets. This multiscale harness records observations, answers, and resource use for joint analysis of evidence acquisition and task outcomes (Secs.~\ref{sec:harness} and~\ref{sec:experiments}).

\item \textbf{Learning to seek evidence.} We combine QA supervision with step-level examples from solver trajectories collected separately on verified training tasks. The resulting \textbf{Sight-35B} improves interactive performance and transfers to external long-video evaluation sets (Secs.~\ref{sec:training_data} and~\ref{sec:utility}).
\end{enumerate}

% Binary criteria refer to documented experiments, not general system limitations.
\begin{table*}[t]
\centering
\small
\caption{From task construction to evaluation and learning: documented capabilities of representative long-video benchmarks and agentic systems. Symbols indicate whether each work meets the stated criterion, not relative performance.}
\label{tab:benchmark_comparison}
\begingroup
\definecolor{cmpGreen}{HTML}{22A65B}
\definecolor{cmpRed}{HTML}{E0524D}
% Color is redundant with shape: check = reported; dash = not reported.
\newcommand{\cmpYes}{\tikz[baseline=-0.5ex]{%
  \fill[cmpGreen] (0,0) circle (0.78ex);
  \draw[white,line width=0.55pt,line cap=round,line join=round]
    (-0.40ex,0)--(-0.10ex,-0.28ex)--(0.43ex,0.32ex);
}}
\newcommand{\cmpNo}{\tikz[baseline=-0.5ex]{%
  \fill[cmpRed] (0,0) circle (0.78ex);
  \draw[white,line width=0.55pt,line cap=round] (-0.36ex,0)--(0.36ex,0);
}}
\renewcommand{\arraystretch}{1.1}
\setlength{\tabcolsep}{5pt}
\begin{tabular*}{\textwidth}{@{\extracolsep{\fill}}lccccc@{}}
\toprule
Benchmark / system &\shortstack{Video-grounded\\task checks} &\shortstack{Adaptive visual\\observation} &\shortstack{Shared interactive\\evaluation} &\shortstack{Observed-frame\\evaluation} &\shortstack{Post-training\\utility}\\
\midrule
Video-MME \citep{fu2024videomme}                &\cmpYes&\cmpNo &\cmpNo &\cmpNo &\cmpNo\\
LongVideoBench \citep{wu2024longvideobench}     &\cmpYes&\cmpNo &\cmpNo &\cmpNo &\cmpNo\\
\addlinespace[3pt]
Active Video Perception \citep{wang2025avp}    &\cmpNo &\cmpYes&\cmpYes&\cmpNo &\cmpNo\\
LongVidSearch \citep{yu2026longvidsearch}       &\cmpYes&\cmpNo &\cmpYes&\cmpNo &\cmpNo\\
TRACE / VES-Bench \citep{liu2026trace}         &\cmpYes&\cmpYes&\cmpNo &\cmpYes&\cmpNo\\
AgentVidBench \citep{an2026agentvidbench}       &\cmpYes&\cmpYes&\cmpYes&\cmpNo &\cmpNo\\
LongVT / VideoSIAH \citep{yang2025longvt}      &\cmpYes&\cmpYes&\cmpNo &\cmpNo &\cmpYes\\
\midrule
\textbf{NexSight (ours)}                       &\cmpYes&\cmpYes&\cmpYes&\cmpYes&\cmpYes\\
\bottomrule
\end{tabular*}
\par\vspace{4pt}
\begin{minipage}{\textwidth}\footnotesize
\cmpYes\ Reported; \cmpNo\ not reported under the criterion.
\textit{Task checks}: candidate QA checked against source videos beyond initial generation/annotation.
\textit{Adaptive visual observation}: agent-selected frames/clips reach a visual solver or observer, excluding caption-only retrieval.
\textit{Shared interactive evaluation}: a common tool interface evaluated across model backbones on the same benchmark.
\textit{Observed-frame evaluation}: actual decoded frames compared with reference evidence, rather than predicted intervals or textual milestones.
\textit{Post-training utility}: training experiments using the work's constructed supervision. NexSight includes SightForge, SightLoop, and SightLine.
\end{minipage}
\endgroup
\end{table*}

\section{Related Work}
\label{sec:related}
\paragraph{Fixed-input long-video understanding.}
Video-MME, LongVideoBench, and LVBench evaluate diverse video content, long-context referring reasoning, and extremely long videos, respectively~\citep{fu2024videomme,wu2024longvideobench,wang2024lvbench}. EgoSchema emphasizes extended temporal reasoning, HourVideo hour-long perception and reasoning, and MLVU diverse long-video tasks~\citep{mangalam2023egoschema,chandrasegaran2024hourvideo,zhou2025mlvu}. MVBench and Perception Test provide skill-level evaluation~\citep{li2024mvbench,patraucean2023perception}, complemented by temporal-detail and cross-time diagnostics~\citep{liu2024tempcompass,cai2024temporalbench,chen2024rextime}. LongVA extends language context to accommodate more video frames~\citep{zhang2025longva}.

\paragraph{Active and agentic video understanding.}
Wang et al.\textquotesingle s VideoAgent, VideoTree, and Active Video Perception develop iterative or query-adaptive video observation~\citep{wang2024videoagent,wang2024videotree,wang2025avp}. Fan et al.\textquotesingle s VideoAgent adds event/object memory~\citep{fan2024videoagent_memory}; T* targets efficient keyframe search~\citep{ye2025temporal_search}. LongVT couples global previews with native video cropping; VideoSIAH supports evidence-sparse QA and tool-integrated training~\citep{yang2025longvt}. Image-based methods offer complementary mechanisms: V* guides visual search, Visual CoT supplies regional supervision, and VLM-R$^3$ trains interleaved region selection and reasoning~\citep{wu2024vstar,shao2024visualcot,jiang2025vlmr3}. ReXTime and VideoEspresso construct video-reasoning supervision~\citep{chen2024rextime,han2025videoespresso}, while Video-R1 and VideoRFT combine supervised reasoning data with reinforcement learning~\citep{feng2025videor1,wang2025videorft}.

\paragraph{Agentic and evidence-grounded video evaluation.}
NExT-GQA and CG-Bench evaluate answer grounding~\citep{xiao2024nextgqa,chen2025cgbench}, while MINERVA diagnoses intermediate reasoning errors~\citep{nagrani2025minerva}. LongVidSearch evaluates multi-hop retrieval planning, while VideoGAIA studies tool-augmented video assistance~\citep{yu2026longvidsearch,zhang2026videogaia}. AgentVidBench evaluates multi-hop video reasoning with tool-agnostic evidence milestones~\citep{an2026agentvidbench}. TRACE / VES-Bench jointly audits decoded-frame coverage and answer correctness for ordering and counting~\citep{liu2026trace}. NexSight connects verifiable construction, raw-visual interaction, and observation-level evaluation (Table~\ref{tab:benchmark_comparison}).

\section{The SightLine Benchmark}
\label{sec:benchmark}
\subsection{Task Definition and Information Boundaries}
We operationalize agentic long-video understanding as task-conditioned observation selection: the model uses the question and accumulated observations to choose subsequent temporal windows or local visual inspections, and decides when to answer within a specified budget.

Given a video $V=\{(x_t,t)\}_{t\in[0,L]}$ and a question $q$, the model selects an observation action $u_k$ from its history $h_k$, receives visual observations $o_k$, or outputs a final answer $\hat a$ and stops:
\begin{equation}
 u_k\sim\pi_\theta(\cdot\mid q,h_k),\qquad
 o_k=\mathcal O(V,u_k),
\end{equation}
where $\mathcal O$ is the frozen video-access protocol. The goal is to acquire and integrate sufficient evidence to answer within budget $B$.

Standard evaluation exposes only the question, permitted video metadata, and access interface; answers, annotated evidence locations, and construction summaries remain hidden. Answers must be grounded in video frames, including visible text, rather than webpages, video titles, external subtitles, or speech transcripts. Evidence Oracle is a separate diagnostic setting. Appendix~\ref{app:repro} specifies the sample and evidence schema.

\subsection{Six Core Capabilities}
\begin{figure}[t]
\centering
\includegraphics[width=\linewidth]{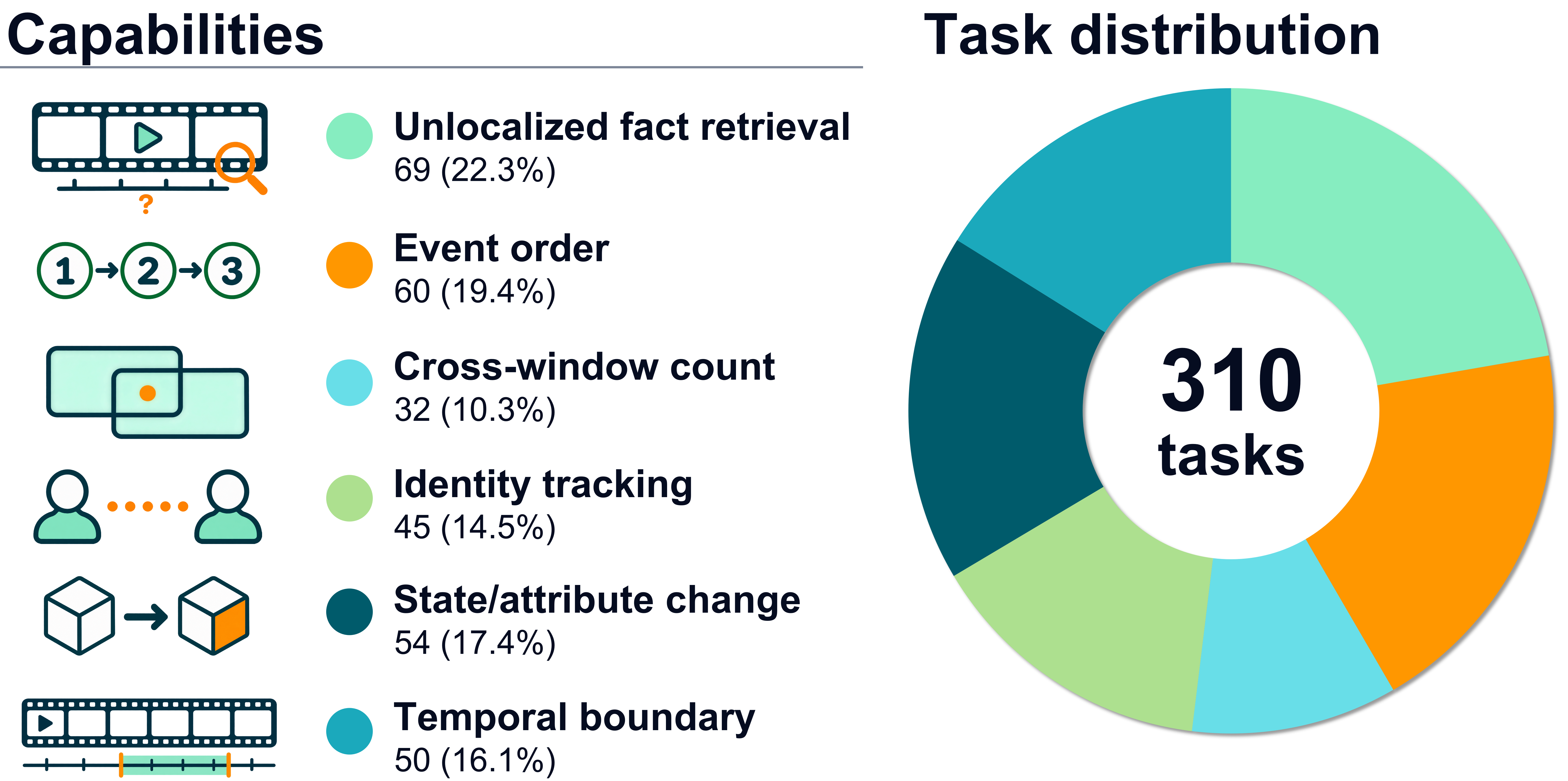}
\caption{\textbf{Six core capabilities and task distribution of SightLine.} The benchmark covers unlocalized fact retrieval without timestamp cues, ordering of separated events, counting across temporal windows without duplication, identity tracking across occlusions and shot changes, state/attribute change recognition, and temporal boundary localization. The donut chart shows the task count and percentage for each capability (310 tasks in total).}
\label{fig:capabilities}
\end{figure}

SightLine covers six fine-grained capabilities (Figure~\ref{fig:capabilities}). Search scope differs from evidence quantity: a single local window may suffice to verify an answer once found, yet locating it can require extensive temporal exploration. Formal identifiers appear in Appendix~\ref{app:repro}.

\subsection{Scale and Composition}
SightLine contains 310 verified tasks from 108 Video-MME Long videos, totaling 74.12 hours across five content domains, with 1,080 annotated evidence artifacts. Figure~\ref{fig:capabilities} and Appendix~\ref{app:distribution} summarize its composition. Original Video-MME questions serve only source-question deduplication and leakage checks, not candidate generation. Appendix~\ref{app:repro} specifies source grouping.

\section{SightLoop: Multiscale Active Observation}
\label{sec:harness}
Task-directed human vision selectively acquires information according to the ongoing task~\citep{hayhoe2005eye}. Motivated by this principle, SightLoop couples two decisions for long-video reasoning: \emph{when to look} along the timeline and \emph{what to inspect} within an observed frame. Its observation loop connects global context, temporal exploration, and local visual inspection with evidence integration.

\subsection{Temporal Exploration and Spatial Inspection}
\textbf{Global overview.} A sparse full-video glance establishes coarse scene and event context. This temporal map guides subsequent exploration.

\textbf{Temporal exploration.} The video agent selects time windows for closer observation and can request earlier, later, or overlapping intervals. Returned frames expose local events that may be absent or ambiguous in the overview. Frame extraction makes selected visual evidence available for subsequent inspection.

\textbf{Local Think-with-Image.} The general-purpose harness provides an image-reasoning sub-agent that can crop, zoom, highlight, and compare regions from extracted frames. These operations let the agent examine answer-critical details such as identity cues, hand--object interactions, text, or state changes, following the image-assisted reasoning paradigm of DeepEyes~\citep{zheng2026deepeyes}. The video agent selects the frame and inspection question; the image agent selects local operations. The image-reasoning sub-agent uses the same evaluated VLM and returns both transformed images and textual findings. It is invoked on demand, with its model calls, visual inputs, and token usage included in the evaluated system's resource accounting.

\subsection{Revisiting and Evidence Integration}
Temporal exploration and spatial inspection form a revisitable observation loop. The agent uses accumulated observations to update its interpretation, identify missing evidence, and choose another window or visual region. For example, determining whether someone took an object before leaving can involve locating the approach and departure, inspecting the hand and tabletop, and revisiting an earlier view to compare object presence.

SightLoop records actual observations, decoded timestamps, and resource use. Optional structured memory (Ledger) retains observation summaries across turns. The agent may answer when it judges the evidence sufficient, or the run terminates at its configured budget. Tool availability and budgets are specified per configuration; Sec.~\ref{sec:matched_observation} isolates temporal selection using fixed window deliveries. Appendix~\ref{app:repro} gives execution and failure-handling details.

\subsection{Answer, Evidence, and Efficiency Evaluation}
We evaluate final answers using the frozen \texttt{gpt-6-luna/high}~\citep{openai2026gpt6luna} semantic rubric and each question's answer specification. Overall accuracy divides correct answers by 310; invalid, missing, and unresolved answers score zero. Temporal evidence coverage is measured by per-question frame-hit recall and its benchmark macro-average:
\begin{equation}
\begin{aligned}
 R_i^{\mathrm{hit}}&=\frac{1}{m_i}\sum_{j=1}^{m_i}
 \mathbb I[T_i\cap I_{ij}\ne\varnothing],\\
 R^{\mathrm{hit}}&=\frac{1}{N}\sum_{i=1}^{N}R_i^{\mathrm{hit}}.
\end{aligned}
\end{equation}
Here $T_i$ contains the decoded timestamps of actually returned frames, $m_i$ is the number of evidence annotations, and $N$ is the number of evaluated questions. Each $I_{ij}$ is the annotated closed interval, with point evidence at $t$ evaluated over $[t-2,t+2]$ seconds. Temporal hits do not imply sufficient visual evidence; answer accuracy and frame-hit recall jointly diagnose evidence-acquisition versus evidence-use failures. We additionally report model/tool calls, observed frames, and recorded token usage to characterize interaction resources. Full scoring and accounting rules appear in Appendix~\ref{app:scoring}.

\section{SightForge: Evidence-Verifiable Data Construction}
\label{sec:construction}
\begin{figure}[t]
\centering
\includegraphics[width=\linewidth]{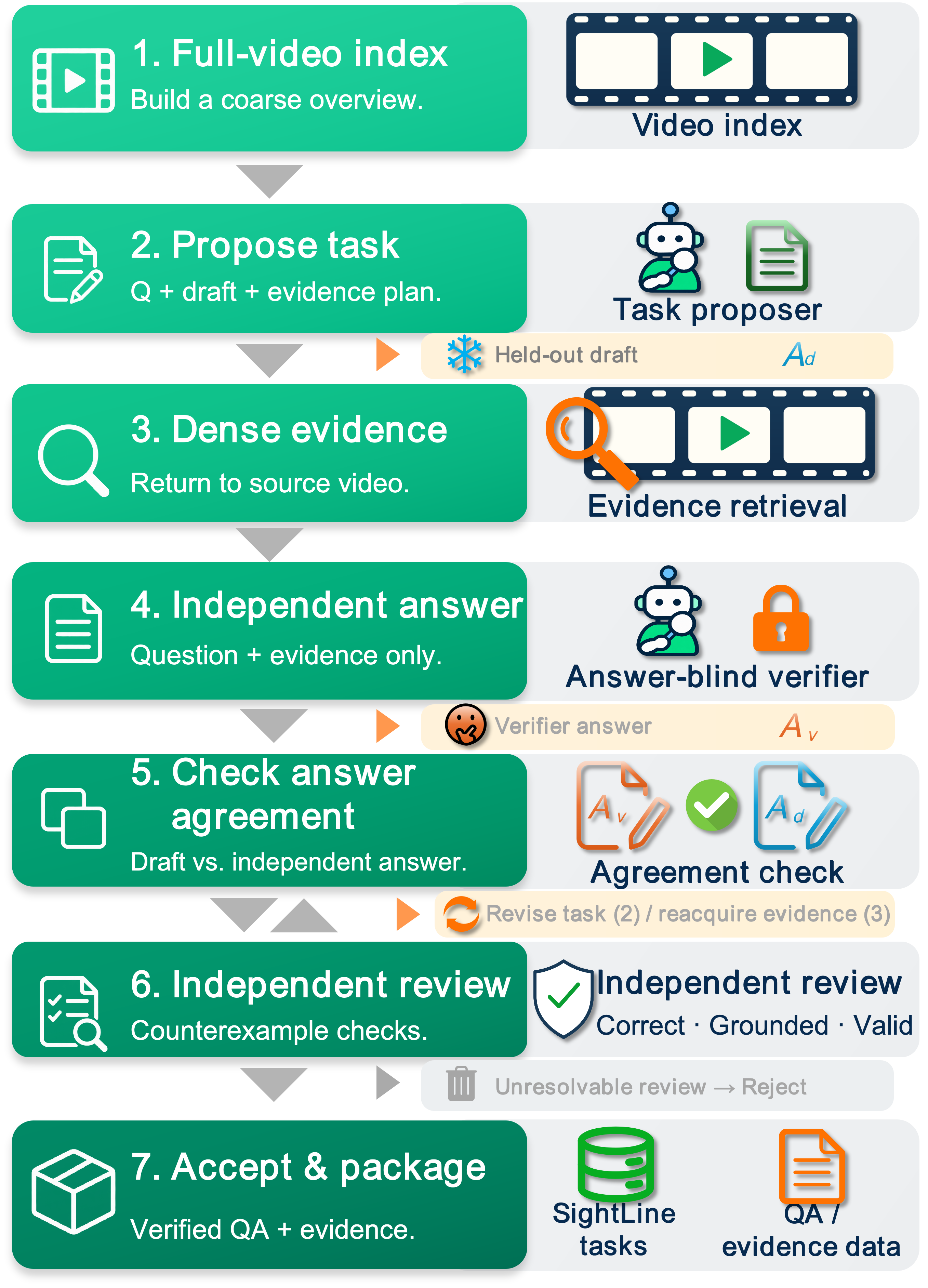}
\caption{\textbf{SightForge: verification before acceptance.} The verifier reconstructs an answer from the question and reacquired evidence, without seeing the draft answer or generation reasoning. Comparison, independent review, and counterexample checks gate acceptance; unresolved candidates return for additional evidence or revision, or are rejected. Solver trajectories are collected separately on training tasks.}
\label{fig:pipeline}
\end{figure}

SightForge uses an agentic skill-based workflow to construct reliable tasks through full-video indexing, candidate generation, evidence acquisition, and verification. Figure~\ref{fig:pipeline} details how candidate tasks acquire raw-video support and pass independent checks before acceptance. The agent schedules these stages under shared skill protocols, yielding verified benchmark tasks or training data according to source-video membership. Construction is strictly separated from the evaluated VLM's task-solving process.

\subsection{Full-Video Indexing and Randomized Observation}
Stratified randomized observations cover the full timeline so that candidate generation does not depend on fixed uniform sampling locations. An observation model extracts visible information from timestamped storyboards; deterministic temporal organization and redundancy compression produce a timeline summary without further free-form model rewriting. The summary guides proposals and search; final evidence is acquired from the raw video.

\subsection{Task Generation and Dense Evidence Acquisition}
Candidate generation proposes a question, draft answer, and evidence plan; verification requires returning to the raw video. The agent reacquires dense evidence from target windows and assembles chronological multi-frame Evidence Panels that preserve local changes and temporal order. Evidence acquisition is task-dependent: boundary tasks retain views before and after a transition; counting tasks cover the specified scope and counting units; identity tasks preserve cross-window identity anchors; and text-related tasks check readability across neighboring frames. These requirements guide dense reacquisition and independent review. The plan also identifies potential counterexamples; gaps or conflicts trigger further observations, revisions, or candidate rejection.

\subsection{Answer-Hidden Verification and Counterexample Checks}
Verification checks \textbf{Answer Correctness} against video facts, \textbf{Evidence Sufficiency} for supporting the answer, and \textbf{Question Validity}: clarity, unambiguity, and answerability. A verifier independently answers from the question and evidence without seeing the draft answer or generation reasoning; its answer is then compared with the draft. Candidates then undergo independent review and counterexample checks.

Independent blind review, conducted by human reviewers or a separately instantiated Codex reviewer, then reconstructs the answer and submits supporting evidence. Counterexample checks inspect adjacent windows and broader task-required ranges; global claims such as counts and first or last occurrences require broader inspection with the search scope recorded. Only candidates passing all three verification dimensions, independent blind review, and counterexample checks enter the task pool for release assembly and validation. Sampling parameters, evidence layouts, and review rules appear in Appendix~\ref{app:repro}.

\subsection{Training Data Construction}
\label{sec:training_data}
Our construction protocol separates evaluation and training by original-source ID ($\mathcal V_{\mathrm{train}}\cap\mathcal V_{\mathrm{test}}=\varnothing$; Appendix~\ref{app:repro}). Beyond static QA/evidence supervision, agentic supervision leverages actual solver interactions collected on training tasks. While construction trajectories record how a question is proposed and verified, solver trajectories record how an agent solves the task from scratch without access to the reference answer or annotated evidence.

Each retained solver call becomes a step-level training example, with its interaction history as context and the current assistant output as the target. Loss is computed only on the current target turn, whereas historical turns, visual inputs, and tool returns serve as context. Tool-call targets retain recorded assistant text, reasoning, and action arguments, while answer targets retain the solver's original responses. Step-level conversion filters out malformed calls or execution errors rather than filtering trajectories by final-answer correctness, preserving diverse exploration behaviors.

\section{Experiments}
\label{sec:experiments}
\label{sec:quality}
\paragraph{Experimental setup.}
All SightLine results and diagnostics use the same 310-task set. Direct-64 and Direct-128 sample uniform fixed frames. We evaluate two complementary interactive settings: the full multiscale SightLoop harness measures end-to-end agentic performance with configuration-specific interaction budgets and optional structured memory (Ledger On/Off); the matched-budget protocol in Sec.~\ref{sec:matched_observation} isolates temporal window selection. SightLine uses the frozen \texttt{gpt-6-luna/high} semantic rubric; external benchmarks use deterministic scoring. Invalid or missing outputs remain in the full denominator. Appendices~\ref{app:repro}--\ref{app:scoring} specify execution and scoring; Table~\ref{tab:harness_budget_profiles} reports recorded full-harness budget profiles.

\subsection{Overall Performance and Passive Context Scaling}
\label{sec:main_results}\label{sec:active}
\begin{figure*}[t]
\centering
\includegraphics[width=\textwidth]{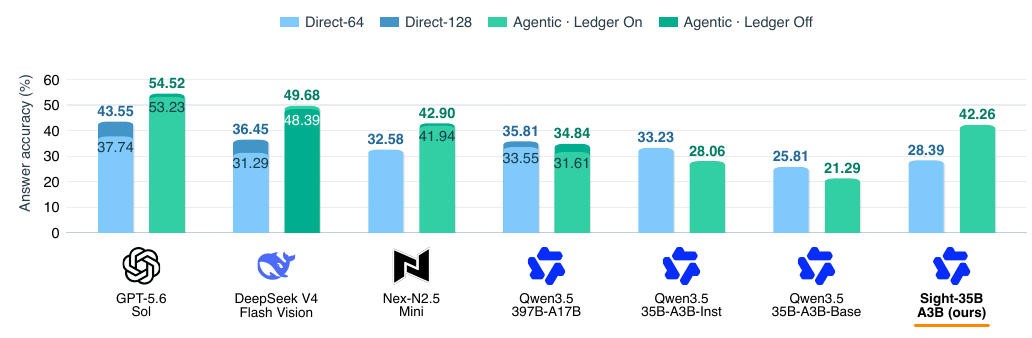}
\caption{\textbf{Overall model performance on SightLine.} Accuracy over 310 tasks across fixed-input Direct and interactive SightLoop configurations (Ledger On/Off). Each model has two overlaid bars: Direct-64/128 and Ledger On/Off. The top and rounded internal boundary mark scores, not their sum. Reference models are ordered by their highest score, followed by Sight-35B. Model sources: GPT-5.6 Sol~\citep{openai2026gpt56sol}, DeepSeek V4 Flash Vision~\citep{deepseek2026v4vision}, Nex-N2.5-Mini~\citep{nexagi2026nex25mini}, Qwen3.5-397B-A17B~\citep{qwen2026qwen35397b}, and Qwen3.5-35B-A3B-Base~\citep{qwen2026qwen35base}.}
\label{fig:overall_results}
\end{figure*}

\paragraph{Frontier models remain challenged despite gains from active exploration.}
Figure~\ref{fig:overall_results} compares frontier and open-weights models under fixed-input Direct and SightLoop configurations across all 310 tasks. GPT-5.6 Sol achieves the highest overall accuracy at 169/310 (54.52\%) with Ledger Off and 165/310 (53.23\%) with Ledger On, followed by DeepSeek V4 Flash Vision at 154/310 (49.68\%) with Ledger On. Under passive Direct evaluation, doubling uniform frame inputs from 64 to 128 consistently improves performance, raising GPT-5.6 Sol from 37.74\% to 43.55\% (+5.81 points), DeepSeek from 31.29\% to 36.45\% (+5.16), and Qwen3.5-397B from 33.55\% to 35.81\% (+2.26). For GPT-5.6 Sol and DeepSeek V4, interactive SightLoop configurations exceed Direct-128, whereas Qwen3.5-397B achieves its highest score under Direct-128.

\begin{figure}[t]
\centering
\includegraphics[width=\linewidth]{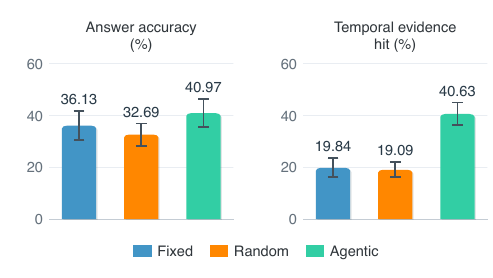}
\caption{\textbf{Temporal selection under matched budgets.} Only window selection varies across a fixed 128-frame budget. Bars show 310-task accuracy and temporal evidence recall ($R^{\mathrm{hit}}$); error bars denote 95\% video-cluster bootstrap confidence intervals. Random averages three seeds per question. The paired Agentic--Fixed temporal-hit difference interval is positive (+20.79 points), while the accuracy-difference interval crosses zero (+4.84 points; Appendix~\ref{app:matched_observation}).}
\label{fig:matched_observation}
\end{figure}

\begin{figure}[t]
\centering
\includegraphics[width=\linewidth]{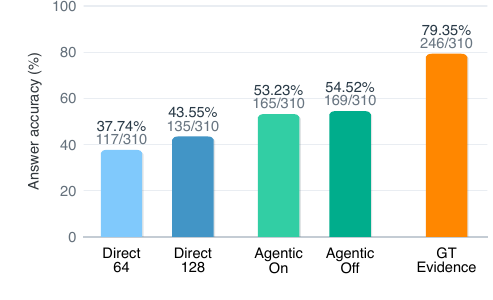}
\caption{\textbf{Evidence-availability diagnostic.} GPT-5.6 Sol performance with actively selected observations versus supplied ground-truth (GT) evidence under the same semantic rubric over all 310 tasks. Evidence Oracle supplies annotated GT evidence artifacts in a single pass with \texttt{xhigh} reasoning. Agentic On/Off denotes SightLoop with Ledger On/Off.}
\label{fig:oracle_diagnostic}\label{fig:evidence_diagnosis}
\end{figure}

\subsection{Matched-Budget Active Observation}
\label{sec:matched_observation}
Because the full SightLoop harness dynamically allocates variable tool calls and spatial inspections, we conduct a strictly controlled matched-budget study to isolate the effect of temporal window selection. Using GPT-5.6 Sol/high on all 310 questions, we compare Fixed-Window Interactive (uniformly spaced windows), Random-Window Interactive (three seeds averaged per question), and Agentic Window Selection. Every trajectory receives an identical budget: 32 overview frames followed by six 30-second windows of 16 frames each (128 delivered frame slots across seven logical calls), with spatial zooming disabled so that only temporal window coordinates vary.

\paragraph{Active search doubles temporal localization; answer gains are smaller.}
Figure~\ref{fig:matched_observation} shows that Agentic selection raises temporal evidence recall ($R^{\mathrm{hit}}$) from 19.84\% (Fixed) and 19.09\% (Random) to 40.63\%. The paired Agentic--Fixed gain is +20.79 points (95\% video-cluster bootstrap CI: $[+16.04,+25.60]$). Agentic accuracy is 40.97\% (127/310), exceeding Random (32.69\%) by +8.28 points (CI: $[+3.81,+12.75]$). The accuracy gain over Fixed (36.13\%, 112/310) is 4.84 points (95\% CI: $[-0.96,+10.51]$) and is not statistically significant. Capability-wise results examine how localization gains relate to answer accuracy (Sec.~\ref{sec:capability}; Appendix~\ref{app:matched_observation}).

\begin{figure*}[t]
\centering
\includegraphics[width=0.65\textwidth]{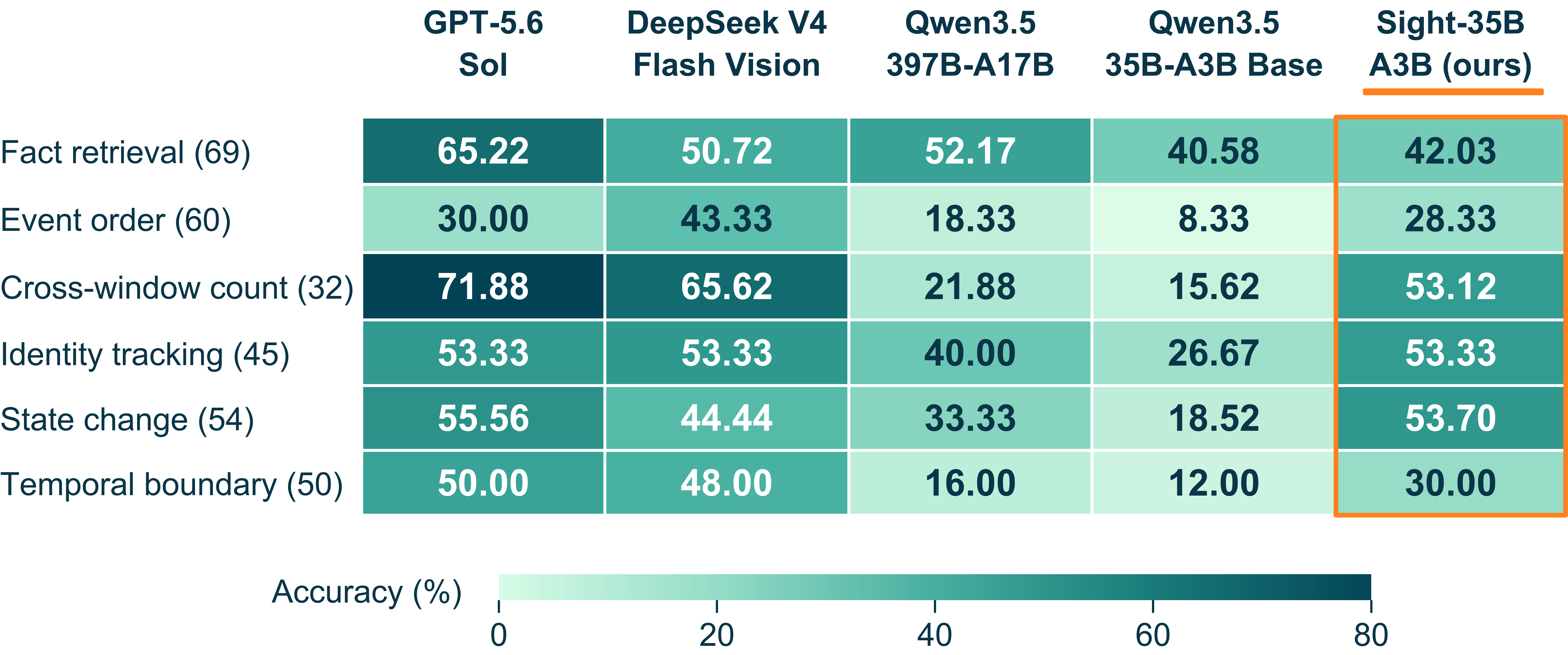}
\caption{SightLoop (Ledger On) capability accuracy (\%) on the 310-task SightLine benchmark; row labels give task counts. Sight-35B gains most over Base in cross-window count and state change under combined post-training, complementing the temporal-localization diagnostic in Sec.~\ref{sec:matched_observation}.}
\label{fig:capability_heatmap}\label{fig:capability_training}
\end{figure*}

\subsection{Evidence Acquisition and Evidence Use}
\label{sec:oracle}
To examine model performance when reference evidence is directly supplied, we evaluate an Evidence Oracle using annotated GT artifacts. GPT-5.6 Sol reaches 246/310 (79.35\%; Figure~\ref{fig:oracle_diagnostic}), versus 53.23\%--54.52\% under SightLoop exploration. Oracle differs in reasoning setting (\texttt{xhigh}), single-call execution, and input format (GT evidence JPGs, including multi-frame panels); together, these conditions probe task performance when annotated visual support is directly available.

Comparing per-question outcomes between standard SightLoop (Ledger On) and the Evidence Oracle yields 145 tasks solved by both, 101 solved only by the Oracle, 20 solved only under standard exploration, and 44 missed by both. These contrasting outcomes guide case-level analysis of observation and answering behavior. Detailed quadrant analyses and qualitative case studies are provided in Appendix~\ref{app:oracle}.

\subsection{Training Utility and Cross-Benchmark Transfer}
\label{sec:utility}\label{sec:transfer}
\paragraph{Agentic post-training substantially improves interactive problem solving.}
We post-train Qwen3.5-35B-A3B-Base~\citep{qwen2026qwen35base} on 5,938 examples constructed via SightForge: 2,623 Direct QA examples and 3,315 step-level interaction examples derived from 837 solver trajectories (comprising 2,499 tool-call steps and 816 final-answer steps). Relative to Base, \textbf{Sight-35B} improves Direct-64 accuracy by 2.58 points and achieves 42.26\% versus 21.29\% (+\textbf{20.97 points}) with SightLoop (Ledger On) (Table~\ref{tab:35b_reversal}).

\paragraph{Fixed-input versus agentic performance.}
The API-served Qwen3.5-35B-A3B-Instruct model leads Sight-35B in Direct-64 (33.23\% vs.\ 28.39\%), whereas Sight-35B leads by \textbf{14.19 points} in interactive evaluation (42.26\% vs.\ 28.06\%; Table~\ref{tab:35b_reversal}). This ranking reversal motivates evaluating passive QA and interactive solving separately.

\begin{table}[t]
\centering\small
\caption{\textbf{Fixed-input and agentic performance of 35B models on SightLine.} Entries report correct answers / accuracy (\%) over 310 tasks; inference budgets vary across deployments.}
\label{tab:35b_reversal}
\setlength{\tabcolsep}{3pt}
\begin{tabular*}{\linewidth}{@{\extracolsep{\fill}}lcc@{}}\toprule
Model / deployment &Direct-64 &Agentic\\\midrule
Qwen3.5-35B-A3B-Base &80 / 25.81 &66 / 21.29\\
Qwen3.5-35B-A3B-Inst &\textbf{103 / 33.23} &87 / 28.06\\
Sight-35B (ours) &88 / 28.39 &\textbf{131 / 42.26}\\\bottomrule
\end{tabular*}
\end{table}

\paragraph{NexSight supervision transfers to external fixed-input benchmarks.}
Under Direct-64, Sight-35B improves over Base by \textbf{8.22 percentage points} on Video-MME Long (900 questions) and \textbf{8.13 points} on the frozen 123-question LVBench subset (Table~\ref{tab:transfer}). See Appendix~\ref{app:transfer_breakdown} for details.

\begin{table}[t]
\centering\footnotesize
\caption{Direct-64 cross-benchmark transfer: correct answers / accuracy (\%), with gains in percentage points.}
\label{tab:transfer}
\setlength{\tabcolsep}{2.5pt}
\begin{tabular}{lcccc}\toprule
Benchmark &Questions &35B Base &Sight-35B &$\Delta$\\\midrule
Video-MME Long &900 &477 / 53.00 &551 / 61.22 &+8.22\\
LVBench subset &123 &55 / 44.72 &65 / 52.85 &+8.13\\\bottomrule
\end{tabular}
\end{table}

\begin{samepage}
\subsection{Capability-Wise Analysis}
\label{sec:capability}
\paragraph{Temporal selection vs.\ cross-window integration.}
The capability breakdown (Figure~\ref{fig:appendix_matched_diagnostics}, Appendix~\ref{app:breakdowns}) shows higher $R^{\mathrm{hit}}$ under Agentic selection across all six capabilities. Accuracy increases for temporal boundary detection (6.00\% $\to$ 36.00\%) and unlocalized fact retrieval (34.78\% $\to$ 44.93\%). Cross-window counting, event ordering, and identity tracking instead show comparable or lower accuracy than Fixed, suggesting that temporal intersection alone does not guarantee sufficient evidence coverage or correct multi-window integration.

\end{samepage}

\paragraph{Capability gains under the combined training recipe.}
Under SightLoop (Ledger On), Sight-35B shows substantial gains on cross-time tasks after combined QA and solver-trajectory post-training (Figure~\ref{fig:capability_heatmap}). Relative to 35B Base, cross-window counting rises from 15.62\% to 53.12\% (+37.50 points), state/attribute change from 18.52\% to 53.70\% (+35.19), identity tracking from 26.67\% to 53.33\% (+26.67), and event ordering from 8.33\% to 28.33\% (+20.00); unlocalized fact retrieval changes modestly from 40.58\% to 42.03\%. The combined recipe improves cross-time task performance and accompanies greater use of explicit spatial inspection: Sight-35B invokes these tools 0.72 times per recorded trajectory versus 0.00 for Base (Table~\ref{tab:tool_breakdown}).

\section{Limitations}
While NexSight advances verifiable evaluation and training for agentic long-video understanding, several limitations remain. First, SightLine focuses on visual evidence and on-screen text across 310 tasks from 108 videos; incorporating audio-visual cross-modal verification and scaling to larger video pools are promising directions for future work. Second, our temporal frame-hit metric ($R^{\mathrm{hit}}$) measures whether decoded frames intersect annotated intervals, which diagnoses temporal localization but does not guarantee that fine-grained spatial details are resolved; combining temporal hit with spatial grounding metrics would further refine trajectory diagnosis. Finally, while our combined QA and trajectory post-training recipe yields substantial gains, future work should isolate the relative contributions of QA supervision and step-level tool imitation under matched training budgets, and investigate whether reinforcement learning provides further gains.

\section{Conclusion}
We presented NexSight for defining, evaluating, and improving fine-grained agentic long-video understanding. SightForge separates candidate generation from raw-video evidence reacquisition, answer-hidden verification, and counterexample checks. SightLine provides 310 verifiable tasks across six capabilities, while SightLoop supports temporal exploration and local inspection under explicit resource accounting. Controlled diagnostics show improved temporal evidence localization, while capability analyses characterize how observation and training gains vary across tasks. Sight-35B, post-trained with combined QA and separately collected solver-trajectory supervision, improves interactive accuracy and external-benchmark scores. Together, checkable task evidence, observation-level evaluation, and targeted supervision provide a foundation for studying and improving active long-video understanding.
\section*{LLM Usage Statement}
\label{sec:llm_usage}
We used \texttt{gpt-5.6-sol/xhigh} for SightForge observation, candidate generation, and answer-hidden verification. Independent Codex-assisted checks complemented the separately recorded human reviews. Training solver trajectories were collected with \texttt{deepseek-v4-flash-vision-exp/xhigh} as the main agent. SightLine answers were scored with \texttt{gpt-6-luna/high} using the frozen semantic rubric. These research uses, information boundaries, and collection settings are specified in Appendices~\ref{app:repro} and~\ref{app:scoring}.

OpenAI Codex assisted manuscript editing, LaTeX preparation, plotting code, and figure/table layout. The authors reviewed the resulting text, code, numerical results, and scientific claims. The author-provided overview and pipeline illustrations were not generated with generative AI; visual evidence frames are extracted from source videos.

{\small
\bibliographystyle{ieeenat_fullname}
\bibliography{references}
}

\clearpage
\appendix
\onecolumn
\raggedbottom
\renewcommand{\thefigure}{A\arabic{figure}}
\renewcommand{\theHfigure}{appendix.\arabic{figure}}
\setcounter{figure}{0}
\FloatBarrier
\section{Supplementary Distributions of SightLine}
\label{app:distribution}
\begin{figure}[!htbp]
\centering
\includegraphics[width=\linewidth]{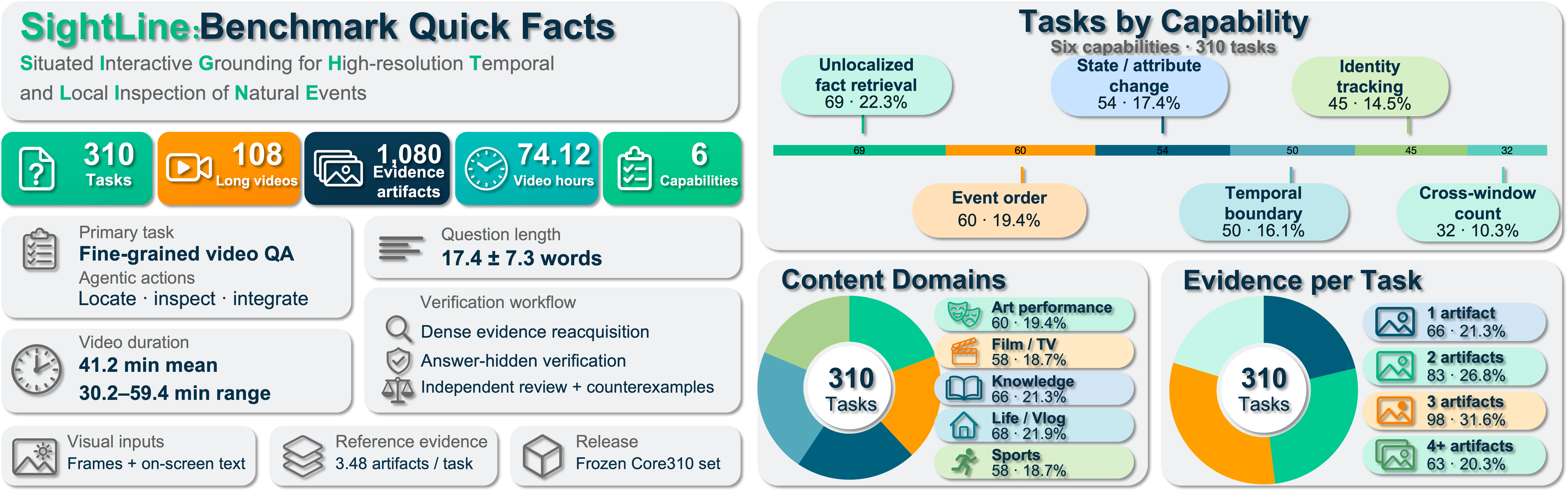}
\caption{\textbf{SightLine benchmark overview and distributions.} The benchmark contains 310 tasks from 108 long videos across six fine-grained capabilities. The figure summarizes benchmark statistics, task proportions by capability and content domain, and the distribution of reference evidence artifacts per task.}
\label{fig:core270_composition}
\end{figure}

SightLine contains 310 tasks across 108 long videos, totaling 74.12 hours. Figure~\ref{fig:core270_composition} and Table~\ref{tab:domain_stats} summarize the content-domain distribution, while Table~\ref{tab:evidence_stats} summarizes overall scale and evidence statistics.

\begin{table}[!htbp]
\centering\small
\caption{Content-domain distribution of SightLine.}
\label{tab:domains}\label{tab:domain_stats}
\setlength{\tabcolsep}{7pt}
\begin{tabular}{lcc}\toprule
Content domain &Videos &Questions\\\midrule
Art Performance &22 &60\\
Film / TV &22 &58\\
Knowledge &21 &66\\
Life / Vlog &22 &68\\
Sports Competition &21 &58\\
\midrule Total &108 &310\\
\bottomrule
\end{tabular}
\end{table}

\begin{table}[!htbp]
\centering\small
\caption{Task scale and evidence statistics. Evidence JPG references denote distinct annotated artifacts (including multi-frame panels).}
\label{tab:stats}\label{tab:evidence_stats}
\begin{tabular}{lc}\toprule
Statistic &Value\\\midrule
Questions / videos &310 / 108\\
Domains / capabilities &5 / 6\\
Total video duration &74.12 h\\
Evidence JPG references &1,080\\
\bottomrule
\end{tabular}
\end{table}

\FloatBarrier
\section{Implementation, Sample Schema, and Reproducibility}
\label{app:repro}
\paragraph{Sample and evidence schema.}
Each sample is defined as $s_i=(V_i,q_i,a_i,\sigma_i,E_i,\mu_i)$, where $a_i$ is the reference answer; $\sigma_i$ specifies answer type, units, tolerance, and allowed aliases; $E_i$ contains evidence intervals and visual artifacts; and $\mu_i$ records content domain, capability, and construction provenance.

Samples record video ID and hash, question ID and text, core capability, content domain, answer specification, evidence intervals, and staged review records. Answer specifications contain kind, value, unit, tolerance, and aliases fixed before evaluation. Evidence items contain start, end, role, description, sampled timestamps, and source hash. Public evaluation inputs are stored separately from answers, evidence locations, and construction summaries.

\paragraph{Capability identifiers.}
The benchmark uses the six formal capability identifiers and task counts listed in Table~\ref{tab:capability_schema}.

\begin{table}[!htbp]
\centering\small
\caption{Formal capability identifiers and task counts.}
\label{tab:capability_schema}
\setlength{\tabcolsep}{4pt}
\begin{tabular}{llc}\toprule
Capability & Identifier & N\\\midrule
Temporal boundary & \texttt{temporal\_boundary} & 50\\
Cross-window count & \texttt{cross\_window\_count} & 32\\
Unlocalized fact retrieval & \texttt{unlocalized\_fact\_retrieval} & 69\\
Identity tracking & \texttt{identity\_tracking} & 45\\
Event order & \texttt{event\_order} & 60\\
State/attribute change & \texttt{state\_attribute\_change} & 54\\
\bottomrule
\end{tabular}
\end{table}

\paragraph{Construction configuration.}
The observation, generator, and verifier stages all use \texttt{gpt-5.6-sol/xhigh}. Full-video indexing samples 72 frames, arranged into six storyboards of 12 frames each, with each frame resized to a width of 320 pixels. Dense evidence is subsequently acquired according to task requirements. The verifier runs in a separate context with the question and reacquired evidence, while the draft answer and generation reasoning are withheld. Review records retain reconstructed answers, answer-hidden status, reviewer type, verification judgments, inspection scope, and rejection reasons.

\paragraph{Stage protocols and full-video indexing.}
Skill protocols specify each stage's input schema, permitted operations, output schema, and acceptance criteria. Videos are selected by domain and subcategory with frozen source versions. Indexing divides the full timeline into 72 segments, selects one randomly jittered frame per segment using a fixed seed, and arranges them chronologically into six timestamped storyboards of 12 frames each. An observation model records visible people, actions, props, scene changes, and on-screen text for each storyboard. A deterministic program organizes observations by timestamp and compresses redundant information into a timeline summary without free-form model rewriting.

\paragraph{Candidate schema and evidence organization.}
Based on the timeline summary, the generator proposes a question, draft answer, task type, structured answer specification, and evidence/retrieval plan. The plan identifies events, time windows, and potential counterexamples to inspect. Evidence organization follows task requirements: boundary questions retain observations before and after a transition; counting questions cover the counting scope and specify units; identity questions retain cross-window identity anchors; and text questions check readability across neighboring frames. The benchmark references 1,080 evidence JPG artifacts. Multi-frame panels preserve local changes and temporal order, enabling reviewers to independently reconstruct answers rather than relying on a single isolated frame.

\paragraph{Independent review and counterexample records.}
Human or independent Codex reviewers reconstruct answers from an answer-hidden review queue and submit supporting evidence, producing judgments of supported, ambiguous, or rejected. Model, Codex, and human reviews are logged separately. Adjacent-window and broader-range checks target missed events, double counting, identity confusion, and shifted boundaries. Global claims such as counts and first or last occurrences require recording the corresponding search scope, combining local evidence support with global counterexample checks.

\begin{figure}[!htbp]
\centering
\includegraphics[width=.60\linewidth]{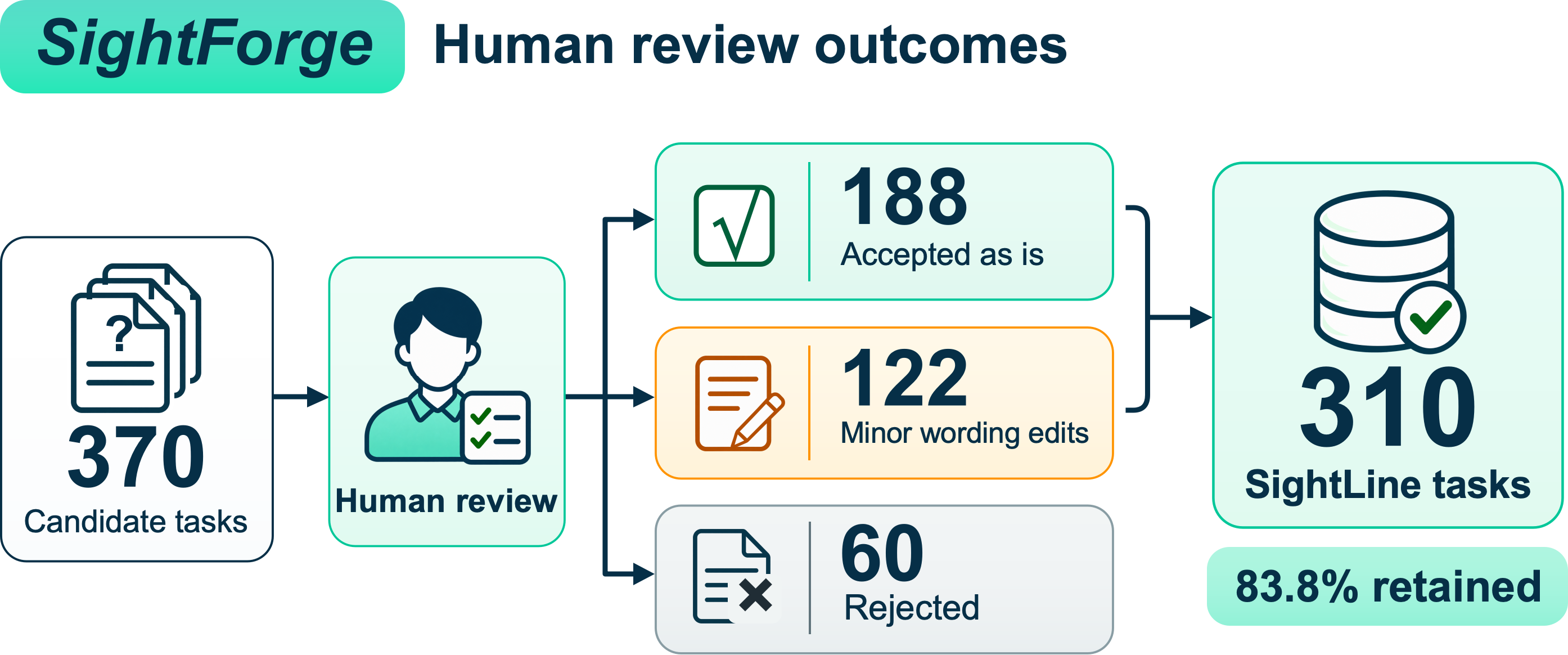}
\caption{\textbf{Human review outcomes for SightForge candidate tasks.} Of 370 candidates, 188 were accepted as is, 122 were retained after minor wording edits, and 60 were rejected. The resulting SightLine benchmark contains 310 tasks, corresponding to an 83.8\% retention rate.}
\label{fig:human_review_outcomes}
\end{figure}

Figure~\ref{fig:human_review_outcomes} summarizes the human review outcomes for the candidate task pool.

\paragraph{Source grouping and training trajectories.}
Source isolation is checked on the video pools and verified on the training records using case-sensitive original-source IDs rather than generated-task IDs. Video-MME numeric IDs are mapped to original video IDs in the selection metadata; LongVILA records are mapped through their source-task manifest. All observations and tasks sharing an original-source ID are assigned to the same source group. The training pool supplies verified QA pairs for static supervision, while agentic supervision collects actual solver interactions on training tasks (question, observation, reasoning, video action, new observation, and final answer). Unlike construction trajectories, solver trajectories operate without access to reference answers or annotated evidence.

\paragraph{Acceptance and packaging.}
Accepted samples pass question-validity, answer-specification, evidence-reconstruction, model blind-verification, independent-review, counterexample, and source-question leakage checks before release assembly. Strict validation covers unique IDs, normalized-question deduplication, video and evidence hashes, timestamp bounds, field validity, and review-record completeness. Release artifacts include the evaluation task file, accepted annotations, a video-selection list, a manifest, and validation records.

\paragraph{Resource budgets and execution rules.}
The resource budget is defined as
\begin{equation}
 B=(T_{\max},F_{\max},P_{\max},K_{\max}),
\end{equation}
constraining interaction turns, actually observed frames, visual pixel resolution, and the input-token limit, respectively. Both the initial overview and subsequent observations count toward the budget; multi-frame panels are charged by their number of source frames. Each interaction logs action parameters, actual returned observations, timestamps, and resource usage, yielding a replayable trajectory. Termination, invalid actions, budget overruns, and technical failures follow a fixed execution protocol. Technical retries cover only transient API failures and never re-run questions based on answer correctness. Accuracy uses the full 310-task denominator, counting invalid or missing answers as incorrect.

\paragraph{Full-SightLoop evaluation configuration.}
Base and Sight-35B use a shared main YAML configuration with the same main prompt and nine-tool interface, Ledger On, \texttt{max\_iterations=100}, a 192k evaluation context, and \texttt{MEDIA\_RESOLUTION\_HIGH}. The YAML specifies limits of 24 model calls, 60 tool calls, 32 \texttt{watch\_clip} calls, one subagent invocation, and 7,200 seconds per task. Actual per-trajectory budget profiles vary across runs and are reported in Table~\ref{tab:harness_budget_profiles} alongside their configuration counts.

\paragraph{Controlled diagnostic configurations.}
Matched temporal selection uses \texttt{gpt-5.6-sol/high}: 32 overview frames plus six 16-frame local windows, 128 delivered frame slots, and seven logical calls, with spatial tools disabled. Evidence Oracle uses \texttt{gpt-5.6-sol/xhigh} with supplied GT artifacts.

\paragraph{Solver trajectory collection.}
Across the 837 retained training trajectories, the root/main-agent requests use \texttt{deepseek-v4-flash-vision-exp} with \texttt{xhigh} reasoning. Collection uses the SightLoop interface for Ledger, video overview, clip observation, frame extraction, OCR/visual experts, and an image subagent. Per-trajectory limits are 14 model calls, 36 tool calls, 24 \texttt{watch\_clip} calls, one subagent invocation, and 1,800 seconds per trajectory, with a 600-second timeout per LLM request. Model-call counts follow the collection-log definition. Solvers receive neither reference answers nor annotated evidence.

\paragraph{Trajectory availability and conversion.}
We collect 837 usable solver trajectories on training tasks, yielding 3,315 step-level training examples (2,499 tool-call steps and 816 final-answer steps). We retain valid step-level interactions without filtering by final-answer correctness to preserve search and recovery behaviors. Together with 2,623 Direct QA examples, these form the 5,938-example training dataset.

\paragraph{Training-example construction.}
Each retained main-agent model call produces one example: interaction history provides context, and only the current assistant output contributes to the training loss. Tool-call targets retain tool names, arguments, and assistant reasoning; answer targets retain the solver's original output. Step-level filters exclude malformed tool arguments and out-of-budget calls.

\paragraph{Training data and source separation.}
Sight-35B is trained on 5,938 examples across 789 original-source video groups: 2,623 Direct QA records from 433 groups and 3,315 trajectory step records from 356 disjoint groups, with zero duplicate lines. Source-ID checks confirm zero overlap between the 108 SightLine evaluation videos and these 789 training sources ($\mathcal V_{\mathrm{train}}\cap\mathcal V_{\mathrm{test}}=\varnothing$).

Sight-35B is fine-tuned from Qwen3.5-35B-A3B-Base~\citep{qwen2026qwen35base} for 186 steps (2 epochs) using seed 1234, sequence length 262,144, global/micro batch sizes 64/1, AdamW with learning rate $2\times10^{-5}$ decaying to $2\times10^{-6}$ via cosine scheduling (10 warmup steps), and weight decay $10^{-4}$. Tensor/pipeline/expert/context parallel sizes are 8/1/8/2. The language model backbone is trainable while the vision encoder and projector remain frozen, with image resolution bounds of 200,704--3,211,264 pixels.

\paragraph{Reproducibility manifests.}
Exact file manifests and SHA-256 checksums for evaluation splits, scoring rubrics, and training records accompany the reproducibility data.

\FloatBarrier
\section{Scoring and Metric Definitions}
\label{app:scoring}
The semantic judge is \texttt{gpt-6-luna/high}~\citep{openai2026gpt6luna}, using a frozen rubric and per-question answer specifications. Judge inputs include the question, reference answer, answer specification, and model prediction; the judge is blind to model identity and does not view the video. Invalid, missing, or unresolved answers score zero. Overall accuracy is $\sum_i c_i/N$ for the evaluated $N$ tasks, and task-macro accuracy is $\frac16\sum_t\sum_{i\in t}c_i/N_t$.

In the response-validity and Oracle tables, \emph{Valid} counts predictions successfully processed by the semantic judge. In the matched-seed table, it counts trajectories with a valid answer, successful execution, and a complete protocol. \emph{Unresolved} counts judgments flagged for review; it can overlap with Valid and is not its complement. Such judgments score zero.

Timestamp-answer tolerances follow the per-question specification (1/2/3/5 seconds). Under a stricter $\min(\text{original tolerance},2\mathrm{s})$ sensitivity check on the matched-budget study, Fixed and Agentic scores remain unchanged at 112/310 and 127/310, while Random changes by a single item (304/930 to 303/930 across three seeds). This answer tolerance is distinct from the $\pm2$-second window used for point-evidence temporal hit ($R^{\mathrm{hit}}$).

$R^{\mathrm{hit}}$ averages the per-question fraction of annotated intervals intersected by actual decoded-frame timestamps. Point annotations use $[t-2,t+2]$ by default, with $\pm1\mathrm{s}$ and $\pm5\mathrm{s}$ evaluated in sensitivity analyses (Figure~\ref{fig:appendix_matched_diagnostics}d). We log logical frame deliveries, unique source frames, multi-turn history image presentations, and API-reported token usage separately. External benchmarks use their standard deterministic scoring rules over full denominators.

\FloatBarrier
\section{SightLoop Implementation Details and Resource Accounting}
\label{app:ablations}
\begin{figure}[!htbp]
\centering
\includegraphics[width=.80\linewidth]{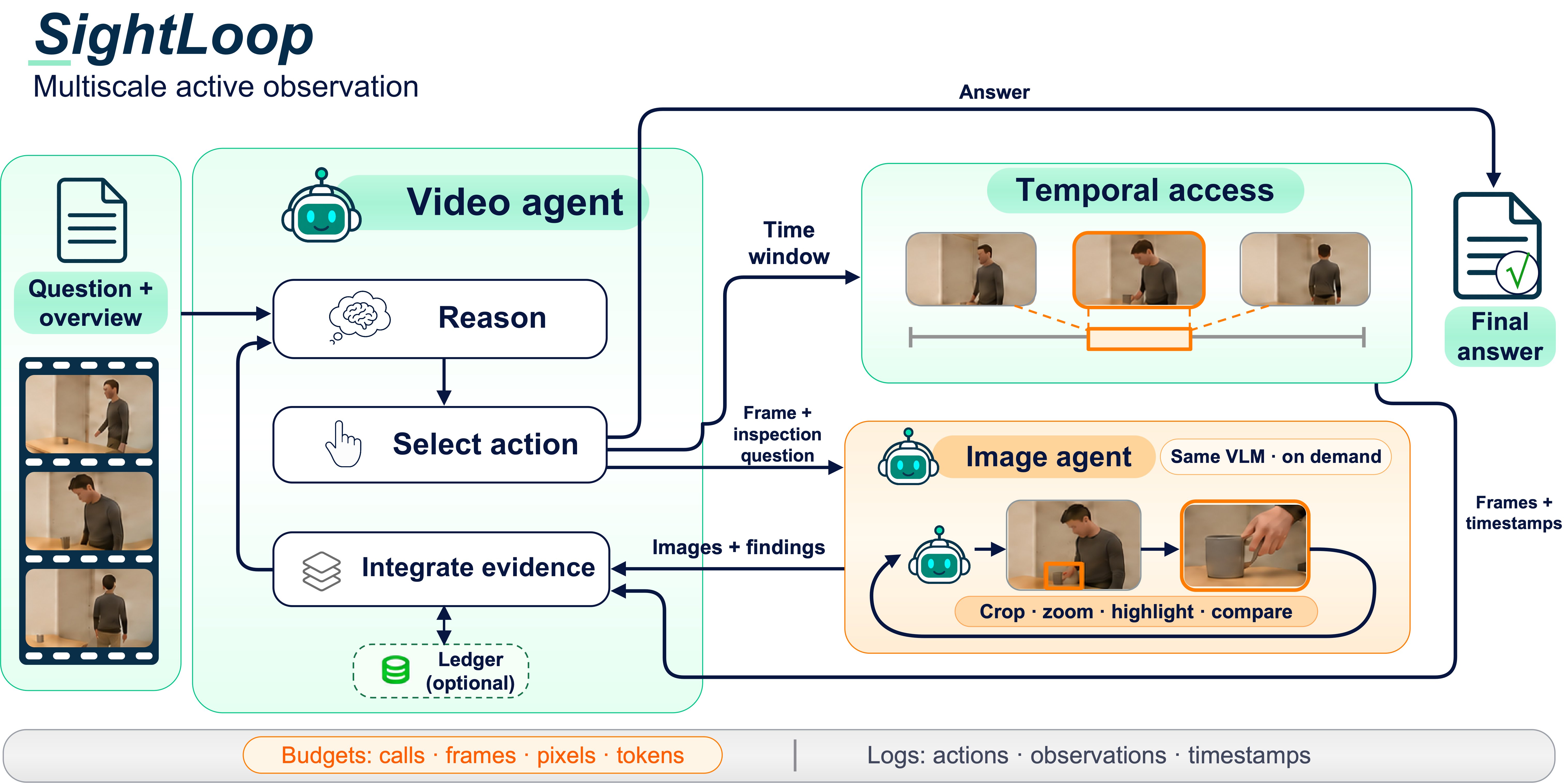}
\caption{\textbf{SightLoop multiscale active observation.} Given a question and video overview, the video agent selects temporal windows, invokes an image agent for local inspection, integrates returned evidence, and revisits observations as needed. The image agent uses the same evaluated VLM. Optional Ledger memory retains observations across turns; calls, frames, pixels, and tokens are governed by explicit budgets, while actions, observations, and timestamps are logged.}
\label{fig:appendix_sightloop}
\end{figure}

Figure~\ref{fig:appendix_sightloop} illustrates the interaction between the video agent, temporal access, and on-demand image inspection.

Tables~\ref{tab:training}--\ref{tab:ledger} summarize model performance, validity rates, and per-question resource consumption across Direct and SightLoop configurations (with structured memory Ledger On/Off). All tables use the same 310-task set. Under Direct-64 evaluation, strict final-answer formatting is produced on 295/310 tasks for the 35B Base model and 224/310 for Sight-35B, with unformatted reasoning-only outputs scored as incorrect in the full denominator.

\begin{table}[!htbp]
\centering\small
\caption{35B checkpoint scores (correct / accuracy \%) over all 310 tasks.}
\label{tab:training}
\setlength{\tabcolsep}{3pt}
\begin{tabular}{lcc}\toprule
Checkpoint &Direct-64 &\shortstack{SightLoop\\Ledger On}\\\midrule
Qwen3.5-35B-A3B-Base &80 / 25.81 &66 / 21.29\\
Sight-35B (ours) &88 / 28.39 &131 / 42.26\\\bottomrule
\end{tabular}
\end{table}

\begin{table*}[!htbp]
\centering\small
\caption{SightLoop accuracy and mean recorded resources per task (310-task normalization).}
\label{tab:main_all}\label{tab:main}
\setlength{\tabcolsep}{5pt}
\begin{tabular}{lccrrrr}\toprule
Model / checkpoint &Ledger &Acc. (\%) &Model calls &Tool calls &Observed frames &Recorded tokens\\\midrule
GPT-5.6 Sol &On &53.23 &11.99 &19.72 &195.99 &331,446\\
GPT-5.6 Sol &Off &54.52 &9.64 &17.82 &211.20 &252,835\\
DeepSeek V4 Flash Vision &On &49.68 &11.26 &14.63 &149.08 &268,386\\
DeepSeek V4 Flash Vision &Off &48.39 &10.16 &13.66 &159.14 &232,136\\
Qwen3.5-397B-A17B &On &31.61 &16.67 &14.06 &274.19 &506,441\\
Qwen3.5-397B-A17B &Off &34.84 &15.08 &12.16 &248.36 &408,774\\
Qwen3.5-35B-A3B-Base &On &21.29 &14.42 &12.51 &180.42 &418,889\\
Sight-35B (ours) &On &42.26 &12.85 &22.82 &219.62 &409,360\\
Nex-N2.5-Mini &On &41.94 &11.78 &19.44 &166.60 &276,343\\
Nex-N2.5-Mini &Off &42.90 &10.45 &16.81 &154.35 &224,801\\
Qwen3.5-35B-A3B-Inst &On &28.06 &12.36 &11.11 &188.99 &379,249\\
\bottomrule
\end{tabular}
\end{table*}

\begin{table*}[!htbp]
\centering\small
\caption{SightLoop response validity and accuracy across all 310 tasks.}
\label{tab:ledger}
\setlength{\tabcolsep}{6pt}
\begin{tabular}{lcccc}\toprule
Model &Ledger &Valid / 310 &Semantic correct &Full-set Acc. (\%)\\\midrule
GPT-5.6 Sol &On &310/310 &165 &53.23\\
GPT-5.6 Sol &Off &310/310 &169 &54.52\\
DeepSeek V4 Flash Vision &On &310/310 &154 &49.68\\
DeepSeek V4 Flash Vision &Off &310/310 &150 &48.39\\
Qwen3.5-397B-A17B &On &303/310 &98 &31.61\\
Qwen3.5-397B-A17B &Off &299/310 &108 &34.84\\
Qwen3.5-35B-A3B-Base &On &310/310 &66 &21.29\\
Nex-N2.5-Mini &On &308/310 &130 &41.94\\
Nex-N2.5-Mini &Off &309/310 &133 &42.90\\
Qwen3.5-35B-A3B-Inst &On &303/310 &87 &28.06\\
Sight-35B (ours) &On &310/310 &131 &42.26\\
\bottomrule
\end{tabular}
\end{table*}

\begin{table*}[!htbp]
\centering\small
\caption{SightLine leaderboard over 310 tasks. Agentic On/Off denotes Ledger On/Off.}
\label{tab:leaderboard310_full}
\begin{tabular}{lccc}\toprule
Model &Configuration &Correct / 310 &Accuracy (\%)\\\midrule
GPT-5.6 Sol &Agentic Off &169 &54.52\\
GPT-5.6 Sol &Agentic On &165 &53.23\\
DeepSeek V4 Flash Vision &Agentic On &154 &49.68\\
DeepSeek V4 Flash Vision &Agentic Off &150 &48.39\\
GPT-5.6 Sol &Direct-128 &135 &43.55\\
Nex-N2.5-Mini &Agentic Off &133 &42.90\\
Sight-35B &Agentic On &131 &42.26\\
Nex-N2.5-Mini &Agentic On &130 &41.94\\
GPT-5.6 Sol &Direct-64 &117 &37.74\\
DeepSeek V4 Flash Vision &Direct-128 &113 &36.45\\
Qwen3.5-397B-A17B &Direct-128 &111 &35.81\\
Qwen3.5-397B-A17B &Agentic Off &108 &34.84\\
Qwen3.5-397B-A17B &Direct-64 &104 &33.55\\
Qwen3.5-35B-A3B-Inst &Direct-64 &103 &33.23\\
Nex-N2.5-Mini &Direct-64 &101 &32.58\\
Qwen3.5-397B-A17B &Agentic On &98 &31.61\\
DeepSeek V4 Flash Vision &Direct-64 &97 &31.29\\
Sight-35B &Direct-64 &88 &28.39\\
Qwen3.5-35B-A3B-Inst &Agentic On &87 &28.06\\
Qwen3.5-35B-A3B-Base &Direct-64 &80 &25.81\\
Qwen3.5-35B-A3B-Base &Agentic On &66 &21.29\\
\bottomrule
\end{tabular}
\end{table*}

\paragraph{Tool-use breakdown.}
Table~\ref{tab:tool_breakdown} summarizes temporal observation, explicit spatial operations, and image-subagent use.

\begin{table*}[t]
\centering\small
\caption{Primary observation-tool calls per recorded SightLoop trajectory ($n$). Spatial operations comprise crop, zoom, highlight, and region comparison. Totals in Table~\ref{tab:main_all} also include memory, metadata, and file tools and use 310-task normalization. Image-subagent invocations are separate from child tool calls; these columns are not additive.}
\label{tab:tool_breakdown}
\setlength{\tabcolsep}{3pt}
\begin{tabular}{lcccccccc}\toprule
Model / deployment &Ledger &$n$ &Overview &\shortstack{Temporal\\window} &\shortstack{Frame\\extract.} &\shortstack{Explicit\\spatial} &\shortstack{Image\\subagent} &\shortstack{Video /\\image OCR}\\\midrule
GPT-5.6 Sol &On &310 &1.04 &11.69 &1.50 &0.25 &0.13 &0.44 / 0.27\\
GPT-5.6 Sol &Off &310 &1.04 &12.33 &1.49 &0.16 &0.10 &0.29 / 0.23\\
DeepSeek V4 Flash Vision &On &310 &1.16 &5.81 &1.55 &0.49 &0.15 &0.70 / 0.28\\
DeepSeek V4 Flash Vision &Off &310 &1.13 &6.91 &1.68 &0.50 &0.17 &0.79 / 0.26\\
Nex-N2.5-Mini &On &308 &1.07 &7.72 &5.05 &0.48 &0.29 &0.56 / 0.49\\
Nex-N2.5-Mini &Off &309 &1.05 &7.76 &4.10 &0.32 &0.23 &0.61 / 0.52\\
Qwen3.5-397B-A17B &On &303 &1.00 &10.76 &0.73 &0.01 &0.07 &0.18 / 0.09\\
Qwen3.5-397B-A17B &Off &299 &1.00 &9.66 &0.59 &0.00 &0.04 &0.13 / 0.06\\
Qwen3.5-35B-A3B-Inst &On &303 &1.00 &7.84 &0.53 &0.00 &0.05 &0.67 / 0.22\\
Qwen3.5-35B-A3B-Base &On &310 &1.00 &9.37 &0.39 &0.00 &0.04 &0.53 / 0.06\\
Sight-35B &On &310 &1.28 &9.32 &3.25 &0.72 &0.14 &1.04 / 0.32\\
\bottomrule\end{tabular}
\end{table*}

\clearpage
\paragraph{Full-SightLoop budget profiles.}
Table~\ref{tab:harness_budget_profiles} summarizes inference-budget profiles in evaluation trajectories matched to the reported predictions by question ID and final answer. Budget allocations vary across models and configurations. Per-question bindings, alternative trace matches, and source-file hashes accompany the reproducibility records.

\begin{table}[!htbp]
\centering\small
\caption{Full-SightLoop evaluation budget profiles. Top: configuration limits. Bottom: counts for one selected trajectory per matched question ($n$), excluding unmatched questions and additional attempts. For 30 configuration--question entries, multiple traces match the prediction; their budget attribution depends on the selected trace. Accuracy retains all 310 questions. Training-trajectory collection and matched temporal selection use separate configurations (Appendix~\ref{app:repro}).}
\label{tab:harness_budget_profiles}
\setlength{\tabcolsep}{4pt}
\begin{tabular}{lcccc}\toprule
Profile &Model calls &Tool calls &\texttt{watch\_clip} calls &Time (min)\\\midrule
B1 &8 &16 &10 &12\\
B2 &10 &22 &14 &18\\
B3 &14 &36 &24 &25\\
B4 &14 &36 &24 &60\\
B5 &14 &36 &24 &120\\
B6 &24 &60 &32 &70\\
B7 &24 &60 &32 &120\\
B8 &24 &60 &32 &150\\
B9 &24 &60 &32 &180\\
\bottomrule\end{tabular}
\par\medskip
\begin{tabular}{lccccccccccc}\toprule
Model / deployment &Ledger &$n$ &B1 &B2 &B3 &B4 &B5 &B6 &B7 &B8 &B9\\\midrule
GPT-5.6 Sol &On &310 &45 &31 &56 &0 &0 &0 &177 &0 &1\\
GPT-5.6 Sol &Off &310 &45 &31 &58 &0 &0 &0 &55 &0 &121\\
DeepSeek V4 Flash Vision &On &310 &45 &30 &56 &155 &24 &0 &0 &0 &0\\
DeepSeek V4 Flash Vision &Off &310 &43 &31 &55 &153 &28 &0 &0 &0 &0\\
Nex-N2.5-Mini &On &308 &41 &20 &45 &0 &0 &0 &202 &0 &0\\
Nex-N2.5-Mini &Off &309 &39 &21 &40 &0 &0 &0 &209 &0 &0\\
Qwen3.5-397B-A17B &On &303 &39 &29 &52 &0 &0 &0 &158 &0 &25\\
Qwen3.5-397B-A17B &Off &299 &42 &31 &51 &0 &0 &0 &146 &0 &29\\
Qwen3.5-35B-A3B-Inst &On &303 &38 &17 &33 &0 &0 &0 &215 &0 &0\\
Qwen3.5-35B-A3B-Base &On &310 &42 &31 &57 &0 &0 &0 &178 &2 &0\\
Sight-35B &On &310 &44 &31 &55 &0 &0 &62 &118 &0 &0\\
\bottomrule\end{tabular}
\end{table}

\clearpage
\section{Full-Set Evidence-Availability Diagnostic}
\label{app:oracle}
Figure~\ref{fig:appendix_oracle_diagnostics} summarizes the Evidence Oracle diagnostic across all 310 tasks. All conditions are scored by \texttt{gpt-6-luna/high} under the same semantic rubric. The Evidence Oracle supplies the ground-truth evidence JPG artifacts (including multi-frame panels) in a single answering call with \texttt{gpt-5.6-sol/xhigh}.

\par\medskip
\noindent\begin{minipage}{\textwidth}
\centering
\includegraphics[width=\textwidth]{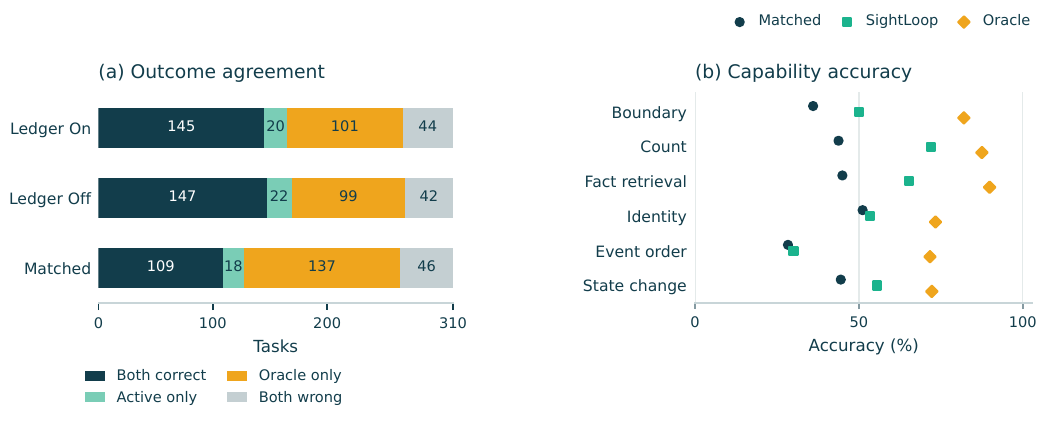}
\captionsetup{hypcap=false}
\captionof{figure}{\textbf{Evidence-availability diagnostic on 310 tasks.}
(a) Per-task outcome counts for each active condition versus the same Oracle.
(b) Capability accuracy for Matched Agentic, SightLoop (Ledger On), and Oracle.
Boundary, Count, Fact retrieval, Identity, and State change abbreviate temporal boundary, cross-window count, unlocalized fact retrieval, identity tracking, and state/attribute change.
Oracle uses supplied GT JPGs and \texttt{xhigh} reasoning in a single pass; protocols and budgets differ, so these are diagnostic rather than controlled causal comparisons.}
\label{fig:appendix_oracle_diagnostics}
\end{minipage}
\par\medskip

Oracle accuracy is 79.35\% (95\% video-cluster CI: [74.68, 83.87]); its difference from Ledger On is +26.13 points (paired 95\% CI: [+19.68, +32.41]). The Oracle solves 101 tasks missed by Ledger On, while both conditions miss 44 tasks. These outcome groups guide case review; assigning a failure cause requires examining the corresponding observations and answers.

The active conditions have 310 judge-processed predictions each, with one unresolved judgment each; Oracle has 309 judge-processed predictions and no unresolved judgments. The active accuracies are 165/310 (53.23\%) for Ledger On, 169/310 (54.52\%) for Ledger Off, and 127/310 (40.97\%) for Matched Agentic. Oracle correct counts in panel~(b), from top to bottom, are 41/50, 28/32, 62/69, 33/45, 43/60, and 39/54. Valid and unresolved counts are not complementary (Appendix~\ref{app:scoring}).

\clearpage
\subsection{Qualitative Examples}
\label{app:qualitative}
\begin{figure*}[!htbp]
\centering
\includegraphics[width=\textwidth]{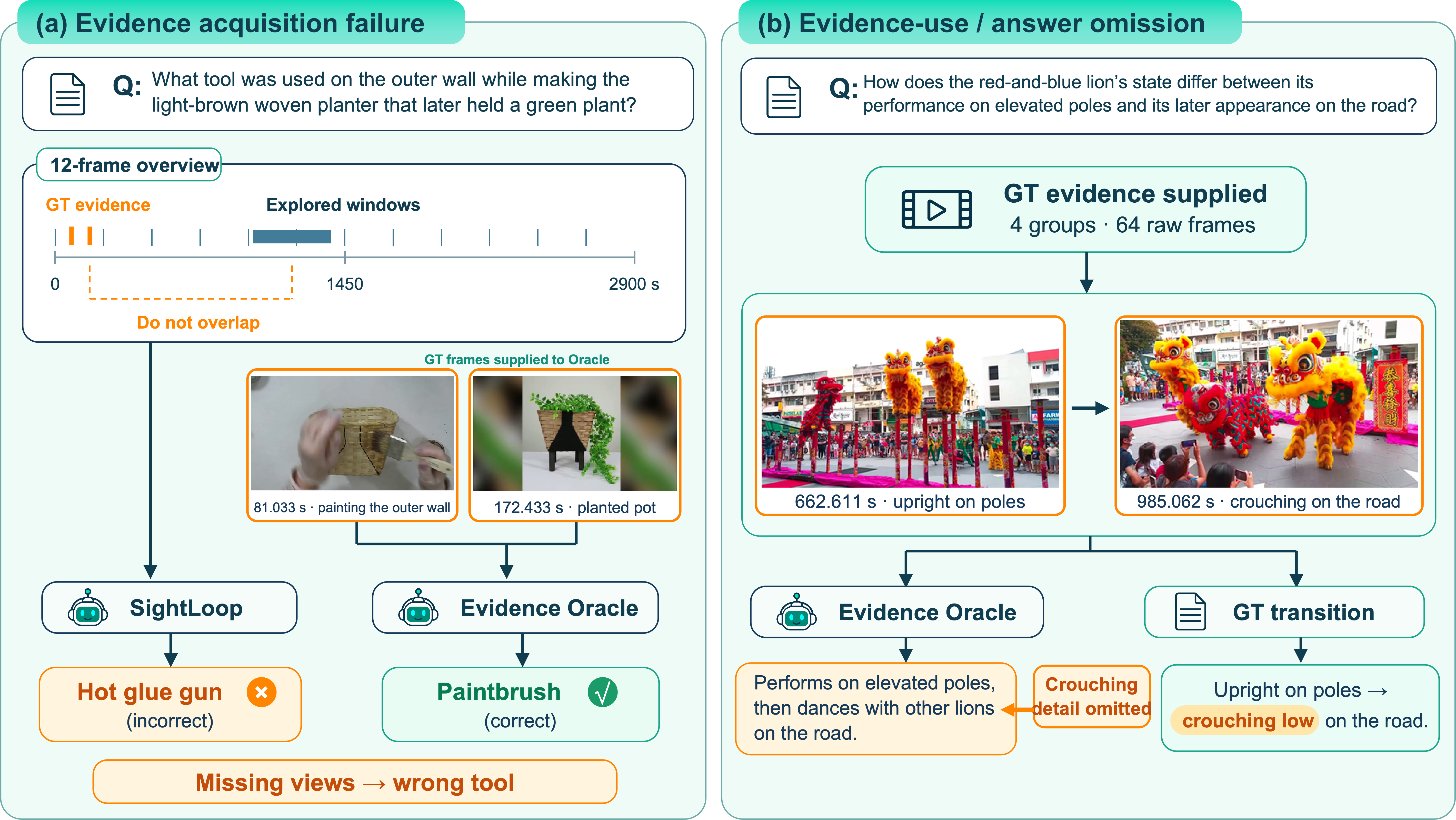}
\caption{\textbf{Qualitative diagnosis of agentic long-video problem solving (GPT-5.6 Sol).} \textbf{Left:} SightLoop timeline ticks indicate returned overview frames, and blue bars show the union of local observation windows. Neither GT evidence group (at $\sim$81\,s and 172\,s) is intersected by the explored windows, whereas supplying GT frames recovers the correct answer. \textbf{Right:} Complete GT evidence is supplied, yet the model response omits the required posture transition.}
\label{fig:qualitative}
\end{figure*}

\begin{figure*}[!htbp]
\centering
\includegraphics[width=\textwidth]{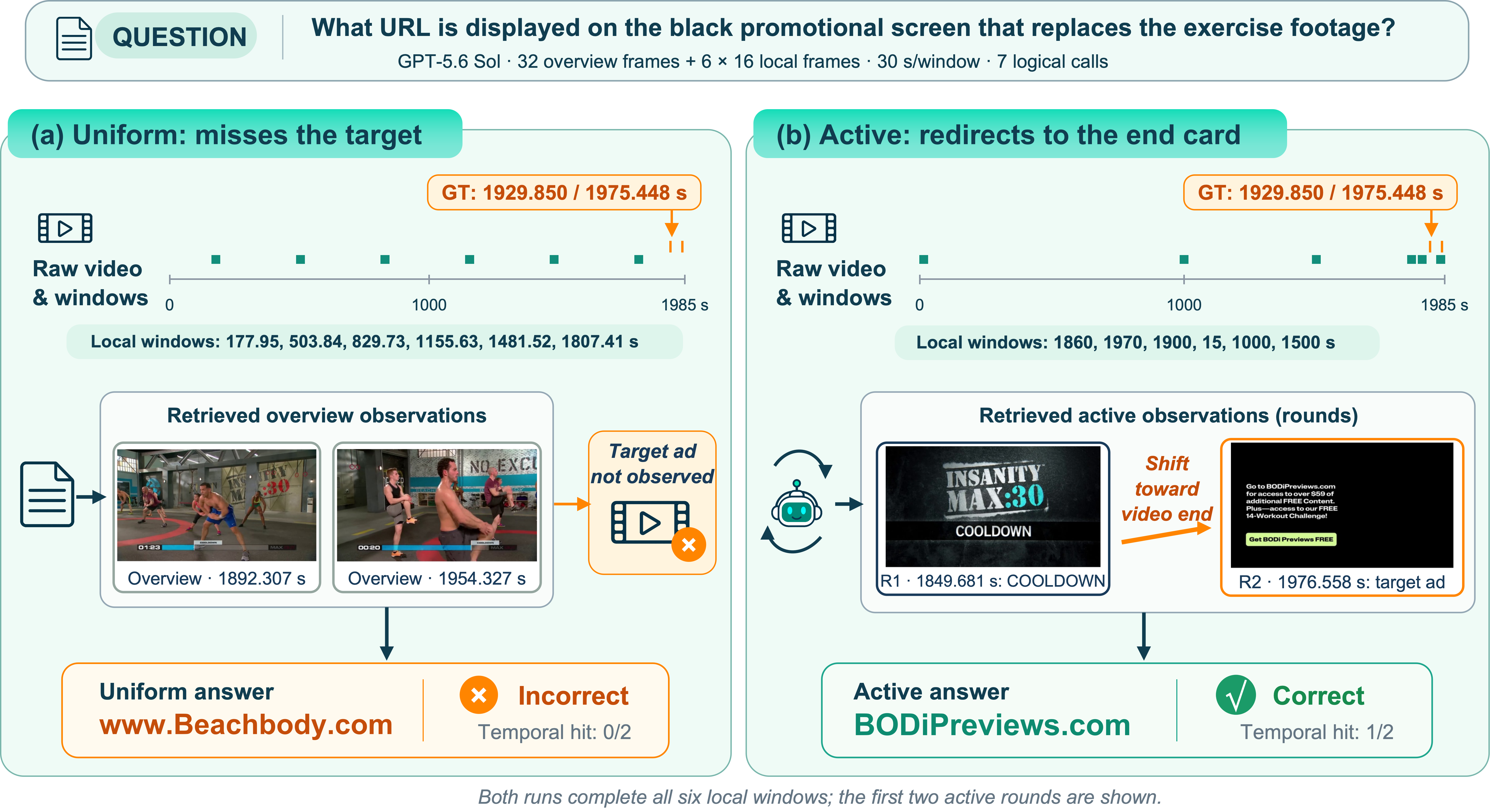}
\caption{\textbf{Qualitative comparison of fixed vs.\ active evidence acquisition (GPT-5.6 Sol).} Under an identical observation budget, Fixed-Window sampling misses both annotated evidence points and answers incorrectly. Active selection redirects subsequent windows toward the end-card region after observing the cooldown segment, capturing the target promotional URL (\texttt{BODiPreviews.com}) and answering correctly ($R_i^{\mathrm{hit}}=1/2$).}
\label{fig:active_success_case}
\end{figure*}

Figures~\ref{fig:qualitative} and~\ref{fig:active_success_case} illustrate representative qualitative failure and success modes. They show evidence-acquisition failure with Oracle recovery, answer-detail omission under supplied evidence, and successful active temporal localization.

\clearpage

\FloatBarrier
\section{Matched-Budget Active Observation: Protocol and Capability Analysis}
\label{app:breakdowns}\label{app:matched_observation}
Figure~\ref{fig:appendix_matched_diagnostics} reports capability-wise accuracy and temporal evidence hit, paired differences, and tolerance sensitivity. Oracle capability results are shown separately in Figure~\ref{fig:appendix_oracle_diagnostics}.

The matched-budget study evaluates 1,550 trajectories across all 310 tasks: 310 for Fixed-Window, $3\times310$ for Random-Window (three random seeds), and 310 for Agentic Window Selection. Every trajectory receives 32 overview frames followed by six 16-frame local observations from 30-second windows (128 delivered frame slots across seven logical calls). Multi-turn history re-presentation yields 560 logical image presentations per trajectory. Random-Window results are averaged across the three seeds per question prior to benchmark aggregation. Confidence intervals (95\%) are computed via 10,000 paired video-cluster bootstrap resamples over the 108 videos.

\par\medskip
\noindent\begin{minipage}{\textwidth}
\centering
\includegraphics[width=\textwidth]{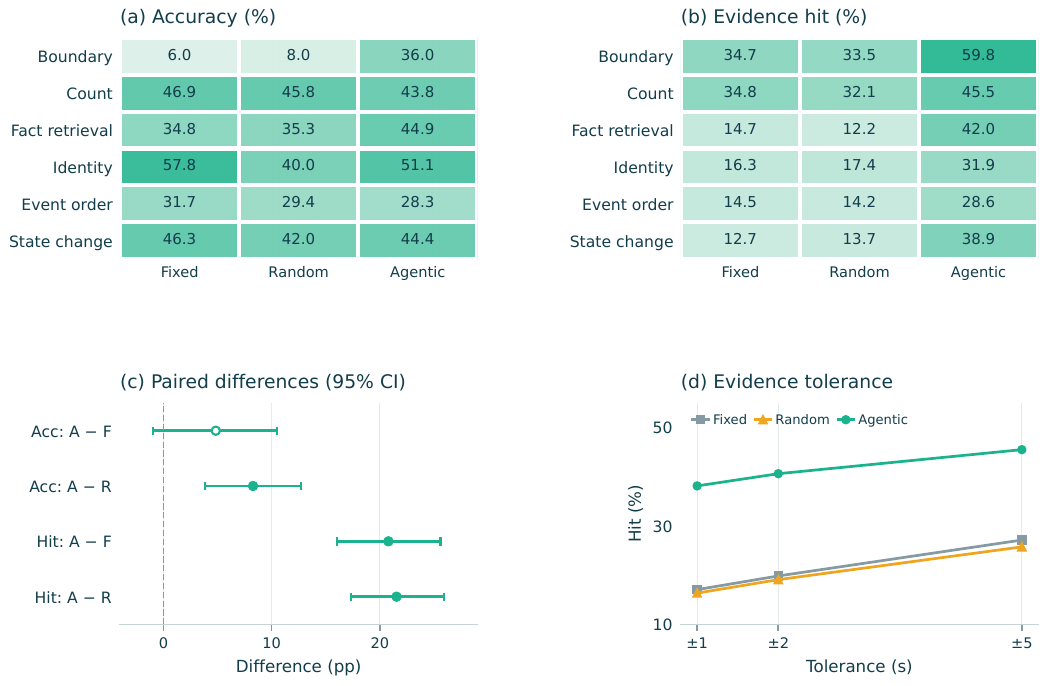}
\captionsetup{hypcap=false}
\captionof{figure}{\textbf{Matched-budget temporal observation diagnostics.}
(a,b) Answer accuracy and temporal evidence hit across six capabilities; both heatmaps use a 0--100\% color scale and display one decimal place. Capability counts in row order are 50, 32, 69, 45, 60, and 54; full capability names follow Figure~\ref{fig:appendix_oracle_diagnostics}.
Random averages three seeds.
(c) Paired video-cluster bootstrap differences with 95\% confidence intervals; A/F/R denote Agentic/Fixed/Random. The Agentic--Fixed accuracy interval includes zero.
(d) Point-evidence tolerance sensitivity. Spatial tools are disabled to isolate temporal window selection.}
\label{fig:appendix_matched_diagnostics}
\end{minipage}
\par\medskip

\begin{table*}[!htbp]
\centering\small
\caption{Per-condition output coverage and seed-level statistics across all 310 questions.}
\label{tab:matched_seed}
\setlength{\tabcolsep}{4pt}
\begin{tabular}{lcccccc}\toprule
Condition & N & Valid & Unresolved & Correct & Acc. (\%) & Hit (\%)\\\midrule
Fixed &310 &309 &0 &112 &36.13 &19.84\\
Random (seed 0) &310 &307 &3 &102 &32.90 &20.38\\
Random (seed 1) &310 &310 &1 &97 &31.29 &17.47\\
Random (seed 2) &310 &310 &2 &105 &33.87 &19.43\\
Agentic &310 &310 &1 &127 &40.97 &40.63\\
\bottomrule
\end{tabular}
\end{table*}

\begin{table*}[!htbp]
\centering\small
\caption{Per-question mean resource accounting in the matched-budget study.}
\label{tab:matched_cost}
\setlength{\tabcolsep}{4pt}
\begin{tabular}{lrrrrcc}\toprule
Condition & Physical calls & Unique frames & API images & Tokens & Repeat (\%) & Overlap (\%)\\\midrule
Fixed &8.14 &127.95 &665.96 &156,160 &0.04 &0.00\\
Random (mean) &8.10 &127.91 &661.97 &155,960 &0.07 &3.14\\
Agentic &8.07 &126.81 &658.43 &155,326 &0.93 &9.00\\
\bottomrule
\end{tabular}
\end{table*}

\noindent\begin{minipage}{\textwidth}
The paired Agentic--Fixed differences are +4.84 percentage points for accuracy (95\% CI: $[-0.96,+10.51]$) and +20.79 for temporal hit ($[+16.04,+25.60]$). Against Random, the corresponding differences are +8.28 ($[+3.81,+12.75]$) and +21.54 ($[+17.36,+25.93]$). Exact two-decimal capability and tolerance values accompany the supplementary data.
\end{minipage}\par

Exact-frame repeats are identified via video identity and source presentation timestamps (PTS), while local-window overlap is defined as one minus the union duration divided by the summed window duration (excluding overview frames). Recorded token means range from 155,326 to 156,160 per question, a spread of 0.53\% relative to Fixed (Table~\ref{tab:matched_cost}).

\FloatBarrier
\section{Cross-Benchmark Transfer Breakdown}
\label{app:transfer_breakdown}
This section supplements Section~\ref{sec:transfer} with detailed breakdowns on Video-MME Long (900 questions) and LVBench (123-question subset). Both benchmarks are evaluated under the Direct-64 setting using standard deterministic multiple-choice extraction over the full question denominators. On Video-MME Long, Sight-35B also exhibits substantially higher generation stability than the 35B Base model, producing valid non-empty final responses on 849/900 questions compared to 729/900 for Base (which experiences more frequent generation timeouts and length overruns). In addition to Direct-64, evaluating LVBench (123-question subset) under an interactive OCR-enabled harness yields a +7.32-point improvement for Sight-35B over the Base model (Table~\ref{tab:transfer_harness_snapshot}).

\begin{table}[!htbp]
\centering\small
\caption{LVBench (123-question subset) evaluation under the interactive harness: correct answers / accuracy (\%).}
\label{tab:transfer_harness_snapshot}
\begin{tabular}{lcccc}\toprule
Benchmark &Questions &35B Base &Sight-35B &$\Delta$\\\midrule
LVBench subset &123 &37 / 30.08 &46 / 37.40 &+7.32\\\bottomrule
\end{tabular}
\end{table}

Figure~\ref{fig:appendix_transfer_breakdown} reports per-category and per-domain accuracies under Direct-64 following each benchmark's native taxonomy.
\par\medskip
\noindent\begin{minipage}{\textwidth}
\centering
\includegraphics[width=\textwidth]{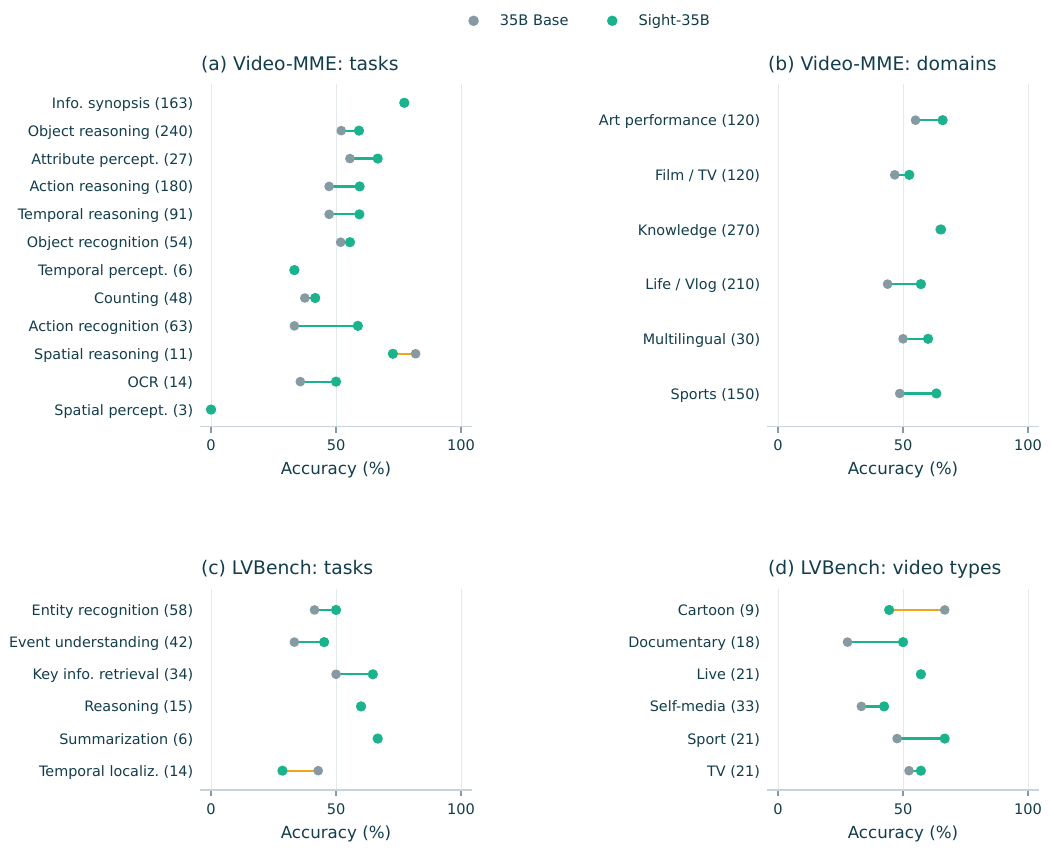}
\captionsetup{hypcap=false}
\captionof{figure}{\textbf{Direct-64 transfer by task and content.}
Gray and mint points show 35B Base and Sight-35B accuracy; orange segments indicate lower Sight-35B accuracy.
Parentheses give category sample sizes.
LVBench task categories can overlap; overall accuracy uses the 123-question denominator.
Video-MME denotes Video-MME Long (900 questions), and LVBench denotes the 123-question subset. Exact category values and full category names accompany the supplementary data.}
\label{fig:appendix_transfer_breakdown}
\end{minipage}
\par\medskip

Across Video-MME Long, Sight-35B achieves consistent gains across major reasoning categories—including object reasoning (+7.08 points), action reasoning (+12.22), temporal reasoning (+12.09), and action recognition (+25.40)—with positive improvements across all six content domains. On LVBench (123-question subset), Sight-35B improves substantially on entity recognition (+8.62), event understanding (+11.90), and key information retrieval (+14.71).

\FloatBarrier
\clearpage
\begingroup
% Source-overlap analysis and closing transparency statements.
\subsection{Source-Video Overlap and Transfer}
\label{app:transfer_sources}
Figure~\ref{fig:source_video_overlap} summarizes source-video overlap among training data, SightLine, and the external evaluation sets. Tables~\ref{tab:videomme_source_transfer} and~\ref{tab:lvbench_source_transfer} compare Base and Sight-35B on the same Direct-64 questions within each source group. Source overlap refers to video identity, not identical training and evaluation questions.

\begin{figure}[!htbp]
\centering
\includegraphics[width=.60\linewidth]{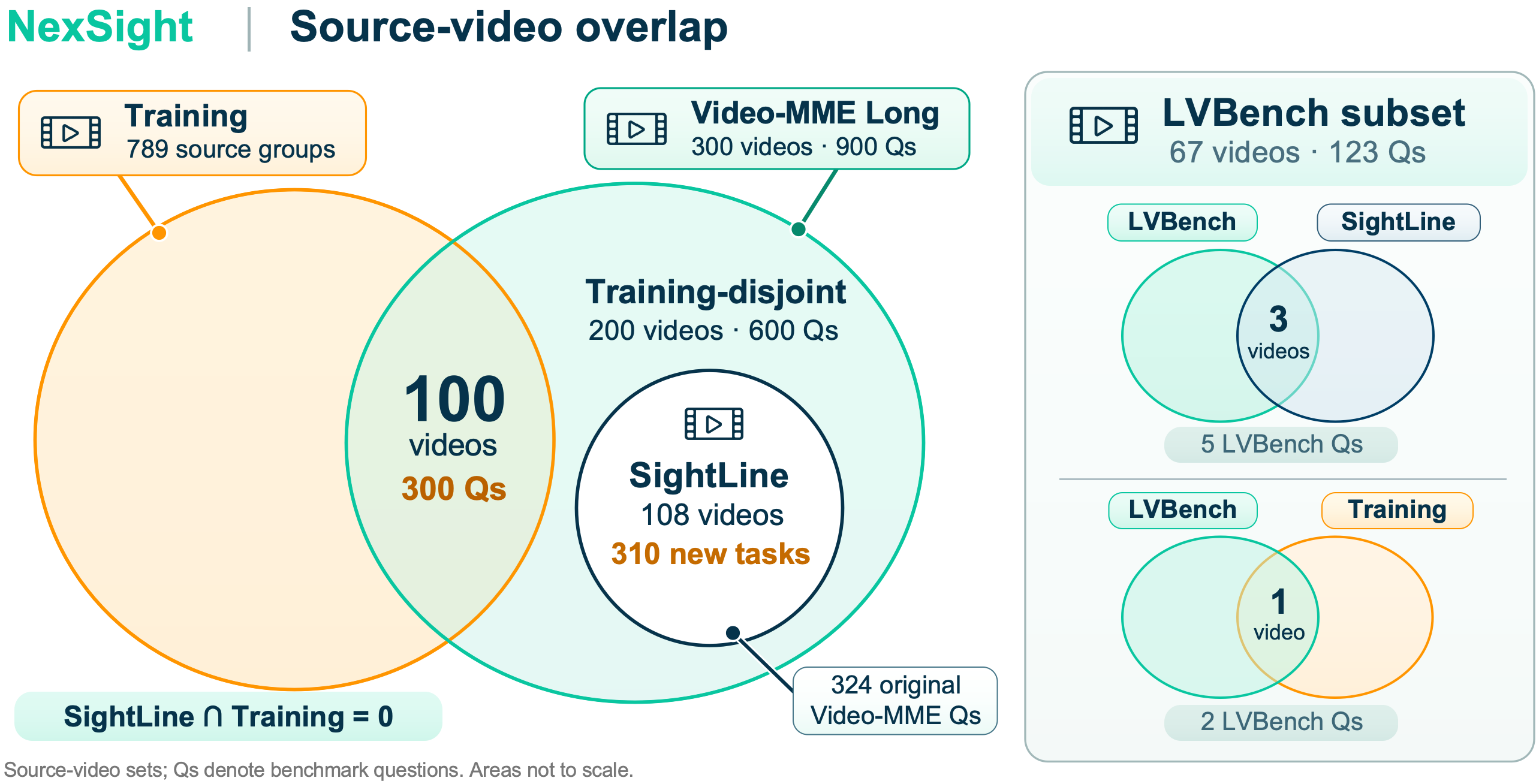}
\caption{\textbf{Source-video overlap across training and evaluation sets.} SightLine contains 310 new tasks from 108 training-disjoint videos, corresponding to 324 original Video-MME questions. Of 300 Video-MME Long videos, 100 overlap with the 789 training sources. The 67-video LVBench subset shares three videos (five questions) with SightLine and one video (two questions) with training. Areas are not to scale; question counts and results appear in Tables~\ref{tab:videomme_source_transfer} and~\ref{tab:lvbench_source_transfer}.}
\label{fig:source_video_overlap}
\end{figure}

\begin{table}[!htbp]
\centering\small
\caption{\textbf{Video-MME Long Direct-64 performance by training-source overlap.} Counts refer to correct answers on the same questions; percentage-point gains are computed from the count difference within each row.}
\label{tab:videomme_source_transfer}
\setlength{\tabcolsep}{7pt}
\begin{tabular}{lccccc}\toprule
Video sources & Questions & Base correct & Sight-35B correct & Net gain & $\Delta$ (pp)\\\midrule
Overlapping training sources & 300 & 173 & 191 & +18 & +6.00\\
Disjoint from training sources & 600 & 304 & 360 & +56 & +9.33\\
\textbf{All} & \textbf{900} & \textbf{477} & \textbf{551} & \textbf{+74} & \textbf{+8.22}\\\bottomrule
\end{tabular}
\end{table}

\begin{table}[!htbp]
\centering\small
\caption{\textbf{LVBench subset Direct-64 performance by training-source overlap.} Both models are evaluated on the same questions within each row. Net gain is the difference in the number of correct answers.}
\label{tab:lvbench_source_transfer}
\setlength{\tabcolsep}{9pt}
\begin{tabular}{lcccc}\toprule
Video sources & Questions & Base correct & Sight-35B correct & Net gain\\\midrule
Overlapping training sources & 2 & 0 & 1 & +1\\
Disjoint from training sources & 121 & 55 & 64 & +9\\
\textbf{All} & \textbf{123} & \textbf{55} & \textbf{65} & \textbf{+10}\\\bottomrule
\end{tabular}
\end{table}

Sight-35B improves in both source groups (Tables~\ref{tab:videomme_source_transfer} and~\ref{tab:lvbench_source_transfer}). Training-source-disjoint questions account for 56 of the 74 additional correct answers on Video-MME Long and nine of ten on LVBench.

\FloatBarrier
\section{Data, Ethics, and Research Transparency}
\noindent\textbf{Data availability.}
Release materials include annotations, evaluation code, evidence/provenance metadata, manifests, and SHA-256 hashes under applicable licenses. Source videos remain under their original platform and dataset licenses and are not redistributed.

\noindent\textbf{Ethics and human verification.}
Public benchmark videos may depict people or copyrighted material. Annotations follow dataset terms without disclosing sensitive personal information. Human blind reviews and automated model/Codex checks are logged separately.

\endgroup
\end{document}